\documentclass[10pt]{article}

\usepackage{booktabs}
\usepackage[utf8]{inputenc}
\usepackage[margin=1in]{geometry} 
\usepackage{multirow}
\usepackage{amsmath,amssymb}
\usepackage{graphicx}
\usepackage{subcaption}
\usepackage{url}
\usepackage{hyperref}
\usepackage{xcolor}
\usepackage{enumitem}
\usepackage{cleveref}
\usepackage{authblk}
\usepackage{mdframed}

\usepackage{hyperref}
\usepackage[colorinlistoftodos]{todonotes}

\title{LUMINA: A Grid Foundation Model for Benchmarking AC Optimal Power Flow Surrogate Learning}

\author[1]{Hongwei Jin}

\author[2]{Keunju Song}

\author[3]{Zeeshan Memon}

\author[1]{Yijiang Li}

\author[4]{Stefano Fenu}

\author[2]{Hongseok Kim}

\author[3]{Liang Zhao}

\author[1]{Kibaek Kim}

\affil[1]{Mathematics and Computer Science Division, Argonne National Laboratory}
\affil[2]{Department of Electronic Engineering, Sogang University}
\affil[3]{Department of Computer Science, Emory University}
\affil[4]{Energy Systems and Infrastructure Assessment Division, Argonne National Laboratory}

\begin{document}

\maketitle

\begin{abstract}
AC optimal power flow (ACOPF) is foundational yet computationally expensive in power grid operations, driving learning-based surrogates for large-scale grid analysis. These surrogates, however, often fail to generalize across network topologies, a critical gap for deployment on grids not seen during training and for routine operational what-if studies. We introduce LUMINA-Bench, a comprehensive benchmark suite for ACOPF surrogate learning covering multi-topology pretraining, transfer, and adaptation. The benchmark evaluates homogeneous and heterogeneous architectures under single- and multi-topology learning settings using unified metrics that capture both predictive accuracy and physics-informed constraint violations. We additionally compare constraint-aware training objectives, including MSE, augmented Lagrangian, and violation-based Lagrangian losses, to characterize accuracy–robustness trade-offs across settings. Data processing, training, and evaluation frameworks are open-sourced as the LUMINA suite to support reproducibility and accelerate future research on feasibility-aware OPF surrogates.
\end{abstract}

\maketitle

\section{Introduction}
\label{sec:intro}

AC optimal power flow (ACOPF) is a foundational optimization primitive in power system operations, used to determine economically optimal generator dispatch while satisfying physical and operational constraints across a network~\cite{carpentier-opf-1962,744492, frank2012optimal_I, frank2012optimal_II}. Two practical challenges limit its use at scale. First, modern grid analysis increasingly demands massive what-if studies, encompassing millions of contingency scenarios in reliability assessments and diverse operating conditions in planning, where ACOPF must be solved repeatedly. Second, ACOPF is nonconvex and nonlinear, typically requiring tailored iterative optimization procedures with prohibitive runtime for large systems. These computational bottlenecks have motivated both hardware-accelerated nonlinear optimization (e.g., \cite{shin2024accelerating,pacaud2025augmented,montoison2025madncl,shin2024scalable,kim2022accelerated,li2024gpu}) and learning-based surrogates (e.g., \cite{pan2022deepopf,huang2021deepopfv,pan2023deepopfal,zeng2024qcqp,fioretto2020predicting,hien2025alternative}).

Learning-based surrogates offer a compelling alternative by amortizing computational cost through offline training, enabling near-instantaneous inference at deployment; more broadly, this direction aligns with emerging efforts on foundation-model perspectives for power grids \cite{hamann2024foundation}. Recent surrogates can be broadly grouped by model class and how they leverage network structure. Deep neural networks (DNNs) learns direct mappings from operating conditions to optimal solutions (or subsets thereof), including DeepOPF~\cite{pan2022deepopf} and DeepOPF-V~\cite{huang2021deepopfv}, with extensions to augmented load conditions~\cite{pan2023deepopfal}. While DNNs can yield strong speedups, graph neural networks (GNNs) have emerged as a natural choice for handling topology and locality, by exploiting the grid's graph structure via message-passing \cite{gao2023physics,liu2022topology,piloto2024canos}. Existing GNN-based surrogates include homogeneous architectures that treat all nodes uniformly~\cite{owerko2020optimal, deihim2024initial, kipf2016semi, xu2018powerful, velivckovic2017graph, yun2019graph} and heterogeneous architectures that explicitly distinguish component types (e.g., buses, generators, loads) \cite{hu2020heterogeneous, mo2022heat, zhang2019hetgnn, busbridge2019relationalgraphattentionnetworks}.

Beyond architecture, constraint handling is central for ACOPF surrogates because feasibility with respect to power flow physics and operational limits is non-negotiable in downstream use. Mean squared error (MSE) captures prediction fidelity to optimal solutions but does not directly enforce feasibility. Augmented Lagrangian (AL) methods~\cite{bouchkati2024augmented} incorporate constraints into training objectives, while violation-based Lagrangian (VBL) methods~\cite{fioretto2020predicting} explicitly penalize constraint violations in the loss. Alternative directions include post-inference correction \cite{wang2024datadrivenacoptimalpower}, loss constructions derived from KKT conditions \cite{nellikkath2022physics,chen2025physics}, and feasibility-oriented architectures for learning constrained ACOPF solutions \cite{zeng2024qcqp,hien2025alternative}. As emphasized in prior work, evaluating ACOPF surrogates require multiple dimensions: predictive accuracy on in-distribution operating points, generalization to unseen operating conditions and topologies \cite{yang2024topology}, scalability across grid sizes, and constraint satisfaction to ensure operational feasibility. However, a systematic benchmark that standardizes tasks, data splits, feasibility metrics, and hyperparameter optimization (HPO) budgets, while evaluating representative baseline architectures and constraint-aware objectives across these dimensions, remains missing.

LUMINA (\textbf{L}arge-scale \textbf{U}nified \textbf{M}odel for \textbf{IN}telligent grid \textbf{A}pplications) is our grid foundation model effort for learning fast, physics-consistent surrogates for ACOPF and related grid optimization primitives. To support systematic and reproducible evaluation, we introduce LUMINA-Bench, a benchmark suite for learning and evaluating ACOPF surrogates across single-topology, multi-topology pretraining, and transfer/adaptation settings. LUMINA-Bench builds on OPFData \cite{lovett2024opfdata}, which provides ACOPF solutions on multiple representative network topologies and diverse operating points. Using standardized protocols, our benchmark suite evaluates a curated baseline set of homogeneous and heterogeneous GNN architectures under matched HPO budgets, and compares widely used constraint-aware training objectives (AL and VBL, as well as MSE). Crucially, LUMINA-Bench reports topology-invariant feasibility metrics that quantify physical validity (e.g., power balance and line-limit violations) alongside predictive error, enabling comparisons across grid sizes and topologies.

\textbf{Contributions.} 
(i) \textbf{Benchmark task suite.} We define standardized benchmark tasks for ACOPF surrogate learning spanning single-topology training, multi-topology pretraining, and transfer/adaptation, with fixed data splits and evaluation protocols based on OPFData \cite{lovett2024opfdata}.
(ii) \textbf{Baseline suite under matched budgets.} We provide a representative baseline set of homogeneous and heterogeneous architectures and compare constraint-aware objectives (AL, VBL) under matched hyperparameter optimization budgets.
(iii) \textbf{Generalization and transfer evidence.} We report results that characterize in-distribution performance, transfer learning, and (where applicable) zero-shot generalization across held-out topologies and larger networks using unified feasibility metrics.
(iv) \textbf{Open-source reproducibility.} We release reproducible data processing, training, and evaluation pipelines as the LUMINA-Bench suite to support research on feasibility aware ACOPF surrogates.

\section{LUMINA-Bench Definition}
\label{sec:background}

This section formalizes LUMINA-Bench by specifying the benchmark tasks, the underlying dataset and topology selection, the ACOPF problem setup, a unified graph I/O schema, and evaluation metrics that jointly capture accuracy and feasibility.


\subsection{Tasks and Protocol}
LUMINA-Bench organizes evaluation into four tasks that stress complementary aspects of ACOPF surrogate learning. 

\textbf{T1: Single-topology training.}
Train and evaluate on a single network topology with disjoint train/validation/test operating points. This task measures in-distribution accuracy and feasibility when the topology is fixed.

\textbf{T2: Multi-topology pretraining.}
Train a single model jointly on multiple topologies and evaluate on the same set of topologies. This task measures whether a shared model can capture common structure across networks and improve sample efficiency relative to training separate models.

\textbf{T3: Held-out topology generalization.}
Train on a subset of topologies and evaluate on an unseen topology that is withheld entirely from training. This task probes topology-shift generalization under the same OPF variable definitions and evaluation metrics.

\textbf{T4: Transfer and adaptation.}
Pretrain a model using one or more source topologies (or smaller networks) and fine-tune using a limited amount of data from a target topology (or larger network). This task evaluates transferability and scaling behavior under matched fine-tuning budgets.

Across all tasks, we report prediction error and topology-normalized feasibility violations defined in Section~\ref{s:metrics}.

\subsection{Dataset and Topology Selection}
\label{subsec:setup:opfdata}

We base LUMINA-Bench on OPFData~\cite{lovett2024opfdata}, which provides solved ACOPF instances across ten representative power network topologies. For each topology, OPFData includes 300K feasible operating points obtained by perturbing load profiles and solving the corresponding ACOPF with a state-of-the-art solver. We use OPFData solutions as supervision and adopt fixed train/validation/test splits per topology to ensure consistent comparisons across tasks, models, and objectives.

Our benchmark uses case30, case57, case118, case500, and case2000. We use case30 primarily for hyperparameter optimization (HPO) and ablations; case57 and case118 for multi-topology training and held-out-topology evaluation; and case500 and case2000 to assess scaling behavior and fine-tuning on larger systems.

We use fixed train/validation/test splits per topology across all experiments; held-out-topology tasks withhold an entire topology (including validation), and transfer/adaptation fine-tunes on a limited subset of the target training split while reporting on the target test split.

\subsection{AC Optimal Power Flow Problem}

For a set of buses $\mathcal{N}$, generators $\mathcal{G}$, and transmission lines $\mathcal{L}$, ACOPF aims to minimize the total generation cost function $C(P_g)$:
\begin{equation}
    C(P_g) = \sum_{g \in \mathcal{G}}(c_{2,g}P_{g}^2 + c_{1,g}P_{g} + c_{0,g}), \label{eq:cost_function}
\end{equation}
where $P_{g}$ is the active power output of generator $g \in \mathcal{G}$, and $c_{2,g}$, $c_{1,g}$, and $c_{0,g}$ are cost coefficients, subject to a set of constraints on power balance, generation, voltage, and line flow as follows.

\emph{Power Balance Equations}: The net active and reactive power injection at a given bus $i \in \mathcal{N}$ must equal the power flow out of the bus into the rest of the network:
\begin{align}
    \sum_{g\in\mathcal{G}_i} P_{g} - P_{d,i} = \sum_{j\in\mathcal{N}_i} P_{ij}(V,\theta), \;
    \sum_{g\in\mathcal{G}_i} Q_{g} - Q_{d,i} = \sum_{j\in\mathcal{N}_i} Q_{ij}(V,\theta), \label{eq:pf1}
\end{align}
where $P_{ij}(V,\theta)$ and $Q_{ij}(V,\theta)$ are the standard AC branch-flow functions induced by the bus admittance matrix $Y_{ij} = G_{ij} + jB_{ij}$.


\emph{Generation Limits}: Active and reactive power generation must be within the physical limits of each generator:
\begin{equation}
    P_{g}^{min} \leq P_{g} \leq P_{g}^{\max}, \;\; Q_{g}^{\min} \leq Q_{g} \leq Q_{g}^{\max} \label{eq:genbounds}
\end{equation}

\emph{Voltage Limits:} Voltage magnitudes and angles must remain within the operational bounds of each bus:
\begin{equation}
 V_i^{\min} \leq V_i \leq V_i^{\max}, \;\; \theta_{i}^{\min} \leq \theta_{i} \leq \theta_{i}^{\max} \label{eq:voltage}
\end{equation}

\emph{Line Flow Limits}: Power flow on a given transmission line must not exceed the line's thermal ratings:
\begin{equation}
    P_{ij}^2 + Q_{ij}^2 \leq \left( S_{ij}^{\max} \right)^2. \label{eq:linelimit}
\end{equation}

Note that power flow equations \eqref{eq:pf1} can be efficiently solved by traditional methods (e.g., Newton-Raphson); however, ACOPF is an NP-hard optimization problem \cite{bienstock2019strong} with inequality constraints \eqref{eq:genbounds}--\eqref{eq:linelimit}.
In LUMINA-Bench, surrogates take operating conditions and topology as input and predict ACOPF decision variables; feasibility is evaluated via the residuals of \eqref{eq:pf1} and \eqref{eq:linelimit}.

\subsection{Graph-Based Input-Output Representation}
\label{s:graphrep}

We cast ACOPF surrogate learning as a graph-based operator learning problem, where the power grid is represented as a heterogeneous graph $G = (\mathcal{V}, \mathcal{E})$ and the surrogate learns a mapping from operating conditions to ACOPF solution variables. 

\textbf{Graph representation.}
The node set comprises four types (bus, generator, load, shunt); edges represent transmission/transformer connectivity with electrical parameters and thermal limits. Full node/edge feature definitions are provided in Appendix~\ref{s:full_feature}.

\textbf{Input schema.}
For a given operating point, the surrogate model receives as input the grid topology and component attributes encoded in the graph representation $G$, including load demand $(P_{d,i},Q_{d,i})$. No ACOPF solution is provided as inputs.

\textbf{Output schema.}
The surrogate predicts a subset of ACOPF solution variables $(V_i,\theta_i,P_g,Q_g)$ for all $i\in\mathcal{N}, g\in\mathcal{G}$. The prediction variables are normalized (via the corresponding min-max bounds) to satisfy the simple bound constraints \eqref{eq:genbounds}--\eqref{eq:voltage}. Accordingly, we report feasibility via residual norms of \eqref{eq:pf1} and \eqref{eq:linelimit}.


\subsection{Evaluation Metrics}
\label{s:metrics}

We evaluate all models using a common metric suite that captures predictive accuracy and physical feasibility.

\textbf{Prediction accuracy.}
We report MSE between predicted and ground-truth ACOPF solution variables, evaluated on both seen and unseen topologies under each benchmark task. We also report accuracy as OPF solution error in tables and figures. 

\textbf{Constraint violation.}
We quantify physical feasibility by measuring violations of each constraint family induced by the prediction. For each constraint type $c \in \{\text{power balance},\, \text{line flow limit}\}$, we compute a per-sample violation vector $v_c^{(s)}$ (stacking bus-wise residuals for power balance or line-wise residuals for flow limits) from the corresponding ACOPF residuals induced by $\hat{\mathbf{y}}$ (power-balance residuals from \eqref{eq:pf1} and line-limit residuals from \eqref{eq:linelimit}). The violation for constraint type $c$ is then defined as the $\ell_2$-norm of the per-sample violation vector, averaged over test samples:
$
    \mathrm{Viol}_c \;=\; \frac{1}{|\mathcal{S}|}\sum_{s\in\mathcal{S}} \left\| v_c^{(s)} \right\|_2 .
$
To enable comparison across network sizes, we report a topology-normalized total violation score defined as the sum of per-type violations normalized by the square root of the number of buses $N$:
$
    \mathrm{Viol} \;=\; \frac{1}{\sqrt{N}}\sum_{c}\mathrm{Viol}_c .
$
We use $\mathrm{Viol}$ as the primary feasibility metric throughout the paper, and additionally report $\mathrm{Viol}_c$ to diagnose which constraint families dominate infeasibility.


%






\section{Baselines and Objectives}
\label{sec:lumina}

We define the baseline suite and training objectives used to validate LUMINA-Bench, covering representative homogeneous and heterogeneous graph architectures and constraint-aware losses under a fixed input/output schema.

\subsection{Baseline Models}
\label{subsec:arch}

We consider eight representative GNN backbones that are commonly used in prior work on learning-based OPF surrogates and cover both homogeneous (single node/edge type) and heterogeneous (typed nodes/edges) message passing. 
All benchmark models share the same graph input/output schema (Section \ref{s:graphrep}) but differ along two axes: (i) backbone architecture and (ii) whether grid heterogeneity (node/edge types) is encoded explicitly. 
All models produce node-level predictions that are assembled into the ACOPF solution vector, are trained under the same objectives (Section \ref{s:losses}), and are evaluated with the same metrics (Section \ref{s:metrics}).
We select widely used backbones that span (i) message passing vs attention and (ii) homogeneous vs typed (heterogeneous) graphs, to cover common OPF-surrogate design choices under a fixed I/O schema.

\textbf{Homogeneous backbones.}
We include three widely used homogeneous message-passing architectures that treat all nodes and edges uniformly: \textbf{GCN}~\cite{kipf2016semi}, \textbf{GAT}~\cite{velivckovic2017graph}, and \textbf{GIN}~\cite{xu2018powerful}. These baselines operate on the graph adjacency and aggregate information from one-hop neighborhoods without explicit node/edge typing. We also include a \textbf{Graph Transformer} baseline that applies multi-head attention over graph neighborhoods (with adjacency-based masking) to capture longer-range interactions beyond standard message passing~\cite{yun2019graph}.

\textbf{Heterogeneous backbones.}
To explicitly represent the heterogeneity of grid components (e.g., buses, generators, loads, shunts) and relational structure, we include four heterogeneous architectures: \textbf{RGAT}~\cite{busbridge2019relationalgraphattentionnetworks}, \textbf{HeteroGNN}~\cite{zhang2019hetgnn}, \textbf{HGT}~\cite{hu2020heterogeneous}, and \textbf{HEAT}~\cite{mo2022heat}. These methods use type-specific projections and/or relation-specific attention to enable different transformation and aggregation rules across node and edge types. In our setting, node types correspond to the component types in Section~\ref{s:graphrep}, and edge types correspond to physical connections (e.g., transmission lines and transformer relations), following the canonical heterogeneous graph formalism.

For completeness, we provide the explicit layer update equations used in our implementations in Appendix~\ref{app:baseline_architectures}.



\subsection{Training Objectives}
\label{s:losses}

Let $\mathbf{y}$ be the ground-truth ACOPF solution labels and $\hat{\mathbf{y}} = f_\theta(G)$ the model prediction. We define residual functions that measure constraint violations induced by $\hat{\mathbf{y}}$: (i) equality residuals $\mathbf{r}(\hat{\mathbf{y}})$ for \eqref{eq:pf1}, and (ii) inequality residuals $\mathbf{h}(\hat{\mathbf{y}}) \leq \mathbf{0}$ for \eqref{eq:linelimit}.

\textbf{Pointwise regression (MSE).}
The default objective minimizes squared error (MSE) on solution variables:
\begin{align*}
    L_{\text{MSE}}(\theta) = \mathbb{E}\left[ \|\hat{\mathbf{y}} - \mathbf{y}\|_2^2 \right].
\end{align*}
This objective measures predictive accuracy but does not explicitly enforce feasibility.

\textbf{Augmented Lagrangian (AL).}
To encourage feasibility during training, we incorporate constraint residuals using an augmented Lagrangian objective \cite{bouchkati2024augmented} as follows:
\begin{align*}
    L_{\text{AL}}(\theta; \boldsymbol{\lambda}, \boldsymbol{\mu}, \rho)
    &= L_{\text{MSE}}(\theta) 
    + \boldsymbol{\lambda}^T \mathbf{r}(\hat{\mathbf{y}})
    + \frac{\rho}{2}\left\|\mathbf{r}(\hat{\mathbf{y}})\right\|_2^2 \notag \\
    &\quad + \boldsymbol{\mu}^T \max\{\mathbf{h}(\hat{\mathbf{y}}), 0\}
    + \frac{\rho}{2}\left\|\mathbf{h}(\hat{\mathbf{y}})\right\|_2^2,
\end{align*}
where $\boldsymbol{\lambda}$ and $\boldsymbol{\mu}$ are dual variables associated with equality constraints and inequality constraints, respectively, and $\rho > 0$ is a penalty parameter.
During the training for $\theta$, $(\boldsymbol{\lambda},\boldsymbol{\mu})$ are updated periodically using ascent steps with projection for nonnegativity:
\begin{equation*}
    \boldsymbol{\lambda} \leftarrow \boldsymbol{\lambda} + \rho\, \mathbf{r}(\hat{\mathbf{y}}),\qquad
    \boldsymbol{\mu} \leftarrow \boldsymbol{\mu} + \rho\, \max\left\{\mathbf{h}(\hat{\mathbf{y}}), 0\right\} .
    \label{eq:al_updates}
\end{equation*}
This objective adaptively reweights constraint satisfaction during training and typically reduces violations under distribution shift.

\textbf{Violation-based Lagrangian (VBL).}
As a feasibility-aware alternative, we use violation-based Lagrangian objective \cite{fioretto2020predicting} as follows:
\begin{align*}
    L_{\text{VBL}}(\theta; \boldsymbol{\lambda}, \boldsymbol{\mu}) 
    &= L_{\text{MSE}}(\theta) 
    + \boldsymbol{\lambda}^T |\mathbf{r}(\hat{\mathbf{y}})|
    + \boldsymbol{\mu}^T \max\left\{\mathbf{h}(\hat{\mathbf{y}}), 0\right\},
\end{align*}
where $\boldsymbol{\lambda}, \boldsymbol{\mu} \geq 0$ are the Lagrangian multipliers that control the accuracy-feasibility trade-off. Similar to AL, the dual variables are updated periodically using ascent steps with step size $\rho>0$:
\begin{equation*}
    \boldsymbol{\lambda} \leftarrow \boldsymbol{\lambda} + \rho\, |\mathbf{r}(\hat{\mathbf{y}})|,\qquad
    \boldsymbol{\mu} \leftarrow \boldsymbol{\mu} + \rho\,\max\left\{\mathbf{h}(\hat{\mathbf{y}}), 0\right\}.
\end{equation*}
This objective adaptively weights constraint violation degrees, as compared to the satisfiability degree \cite{fioretto2020predicting}.

\section{Experimental Protocol}
\label{sec:setup}

This section describes the experimental protocol across all experiments in this paper.

\subsection{Training and Hyperparameter Optimization}
\label{subsec:setup:training}

We employ a unified HPO and training protocol across the eight architectures to ensure fair comparison. All HPO is conducted on case30 using a fixed training budget of 2M samples seen per run. We use W\&B Sweep with Bayesian method and hyperband-based early stopping, with a minimum resource of 600k samples-seen for pruning underperforming trials.

\textbf{Model selection criterion.} Each trial is evaluated on a validation set using a validation score that combines the prediction MSE and the total constraint violation, where the violation term is normalized by $\sqrt{\mathcal{N}}$ to reduce sensitivity to network size. At the end of each HPO stage, we select the configuration with the best (lowest) validation score ($= \text{MSE}_\text{val} + \text{Viol}_\text{total,val}/\sqrt{\mathcal{N}}$).

\textbf{Two-stage HPO.} We propose a two-stage HPO. In stage 1, we tune architecture- and optimizer-related hyperparameters over a shared search space including learning rate, weight decay, dropout, hidden dimension, number of layers, batch size, and gradient clipping norm, using 50 trials per architecture.
In Stage 2, we fix the best Stage-1 configuration each from the homogeneous and heterogeneous architecture groups and then sweep hyperparameters specific to Lagrangian-based objectives, covering both the augmented Lagrangian (AL) and violation-based Lagrangian losses.

Across all runs and both stages, we train with AdamW and apply gradient clipping using the tuned clipping norm.
We leave the full hyperparameter settings and discuss observations during HPO in Appendix~\ref{app:hyperparams}.

\subsection{Implementation Invariants}

To isolate the effects of backbone architecture and training objective, we standardize all other implementation choices across baselines.

\textbf{Shared I/O and head.} All models use the same heterogeneous graph schema (Section~\ref{s:graphrep}) and a shared output head that maps final node embeddings to predicted ACOPF variables, including identical normalization and clipping for box constraints.

\textbf{Matched training budgets.} We train all backbones under the same samples-seen budget and early-stopping logic, and we apply the same two-stage HPO protocol with a matched number of trials per model family (Section~\ref{subsec:setup:training}).

\textbf{Optimization and regularization.} We use the same optimizer family (AdamW), gradient clipping, and dropout conventions across models; only architecture-specific hyperparameters (e.g., number of layers, hidden dimension, attention heads) vary within the HPO search space.

\textbf{Metric reporting.} We report the same accuracy and feasibility metrics for all models (Section~\ref{s:metrics}), using fixed dataset splits and consistent aggregation over random seeds.








\section{Benchmark Results and Analysis}
\label{sec:results}

This section reports benchmark results across tasks and distills empirical findings on how architecture, training regime (single vs.\ multi-topology), and constraint-aware objectives affect accuracy, feasibility, and transfer behavior.


\subsection{Single- and Multi-Topology Baselines}
\label{subsec:pretrain_baseline}



\begin{figure}[!htbp]
    \centering
    \begin{subfigure}[b]{\columnwidth}
        \centering
        \includegraphics[width=0.6\linewidth]{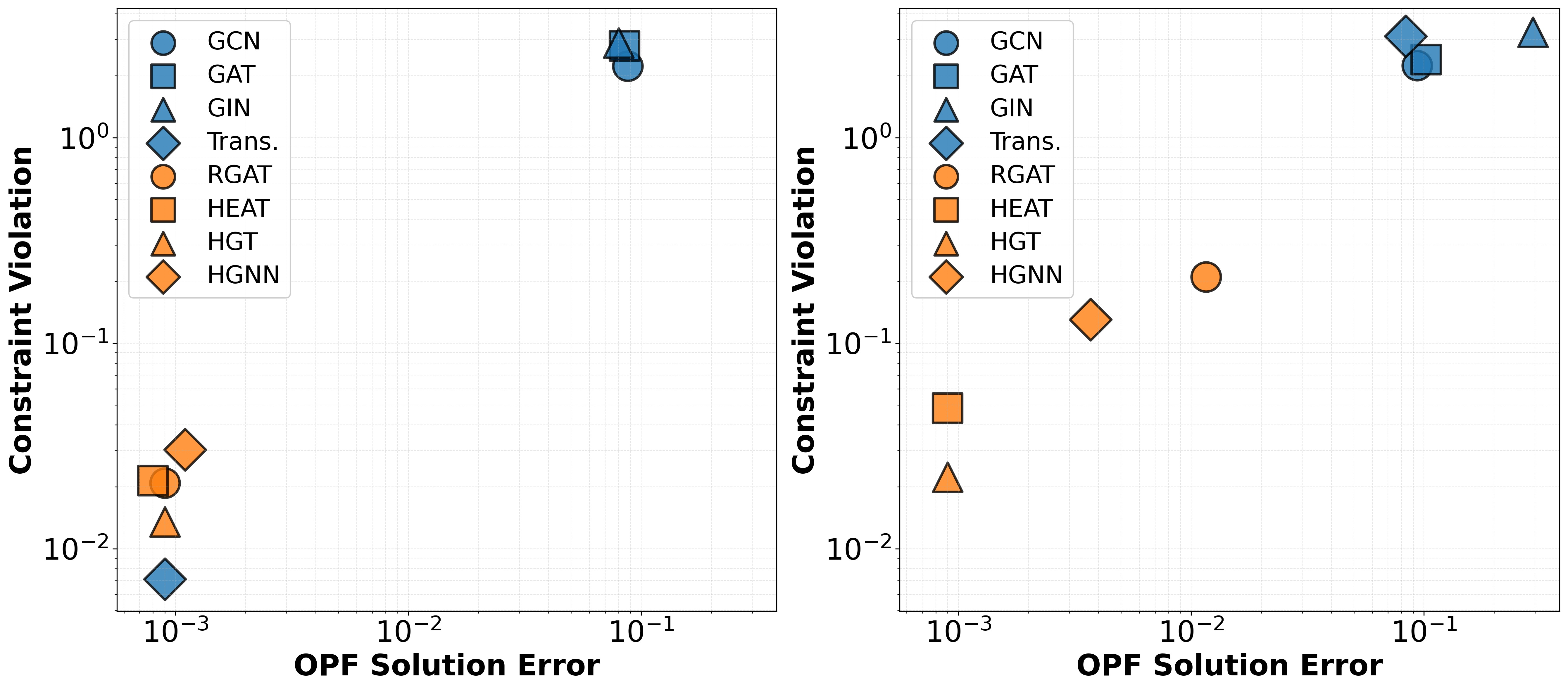}
        \caption{Evaluation on case30}
        \label{fig:case30_eval}
    \end{subfigure}
    
    \begin{subfigure}[b]{\columnwidth}
        \centering
        \includegraphics[width=0.6\linewidth]{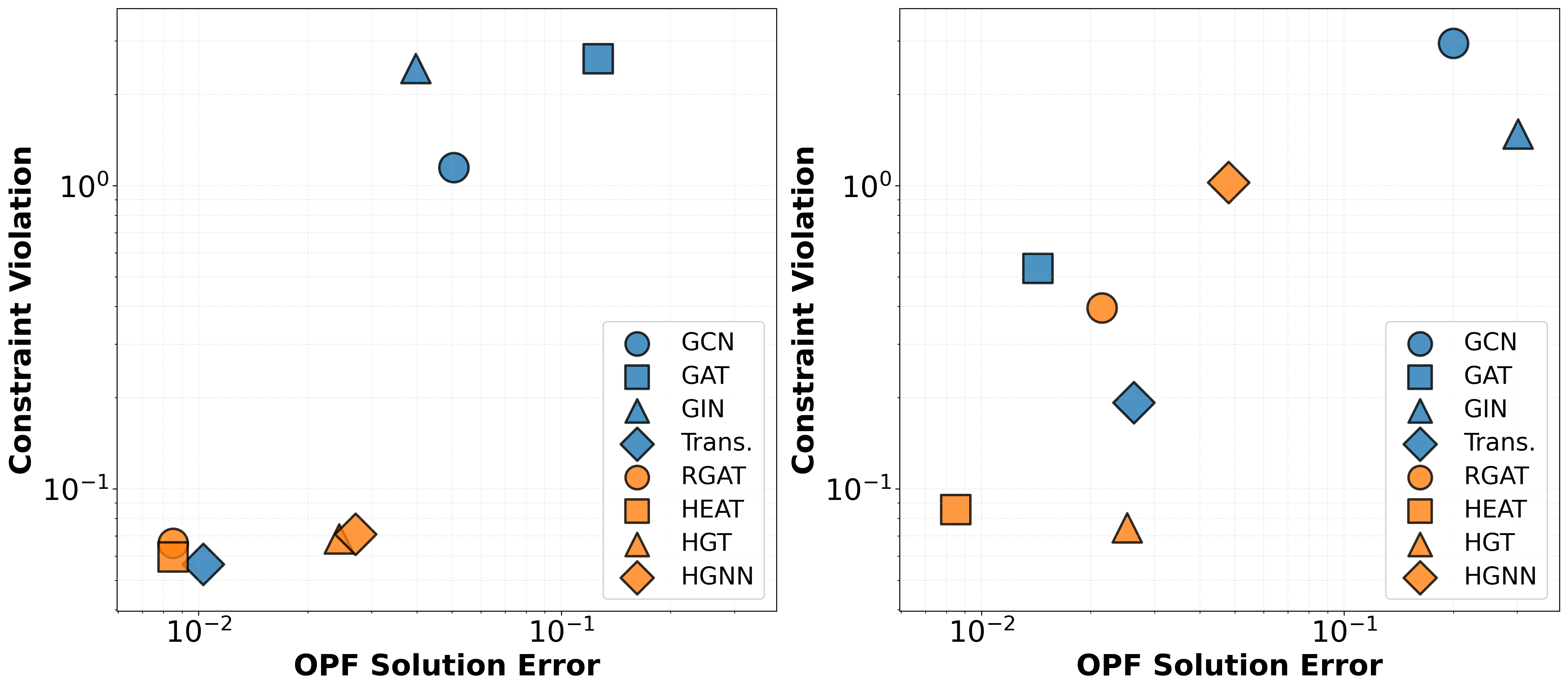}
        \caption{Evaluation on case57}
        \label{fig:case57_eval}
    \end{subfigure}
    
    \begin{subfigure}[b]{\columnwidth}
        \centering
        \includegraphics[width=0.6\linewidth]{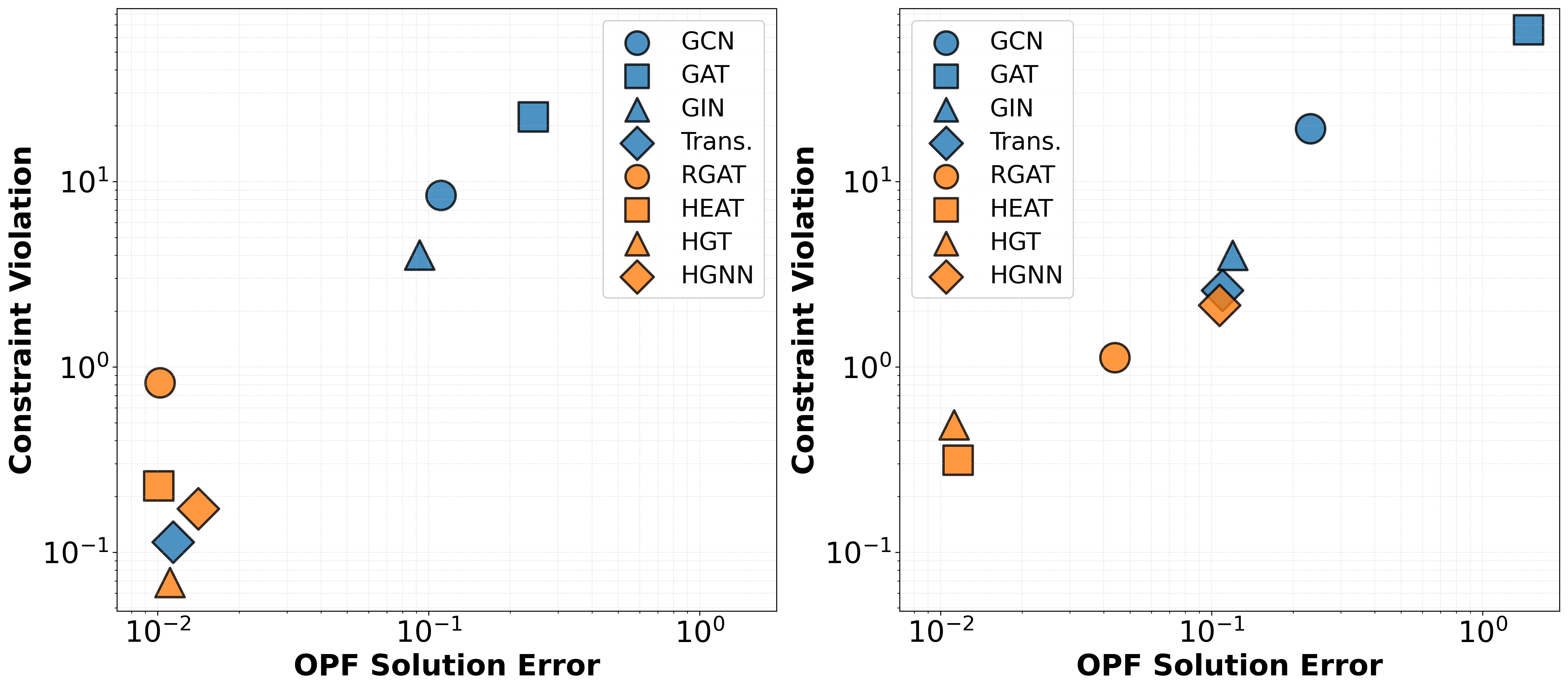}
        \caption{Evaluation on case118}
        \label{fig:case118_eval}
    \end{subfigure}
    
    \caption{Architecture comparison on single-topology (left) vs. multi-topology (right) training. Heterogeneous architectures generally demonstrate better performances than homogeneous models in both solution quality and constraint satisfaction, especially in multi-topology training.}
    \label{fig:topology_evaluation}
\end{figure}

Figure~\ref{fig:topology_evaluation} shows a consistent architecture ranking across all three test topologies. With single-topology training (left panels), heterogeneous models cluster in the lower-left region, achieving both lower solution error and smaller violations; among homogeneous baselines, transformer is the only one that remains competitive, while GCN/GAT/GIN show significantly higher errors and violations. With multi-topology training (right panels), heterogeneous models largely retain their advantage with only modest degradation, whereas homogeneous models deteriorate more noticeable in both accuracy and feasibility. HGT and HEAT are the most robust across both regimes. Violations increase under multi-topology training, particularly evident on case118, consistent with reduced per-topology samples in joint training, but remain within acceptable bounds for heterogeneous models, indicating effective cross-topology knowledge transfer. As network size gwos, both prediction error and constraint violations increase, with feasibility degradation more pronounced on larger cases, especially under multi-topology training. Detailed results are provided in Appendix~\ref{app:additional_experiments}.

\subsection{Comparison of Graph Architectures}

\begin{figure}[!htbp]
    \centering
    \includegraphics[width=0.5\linewidth]{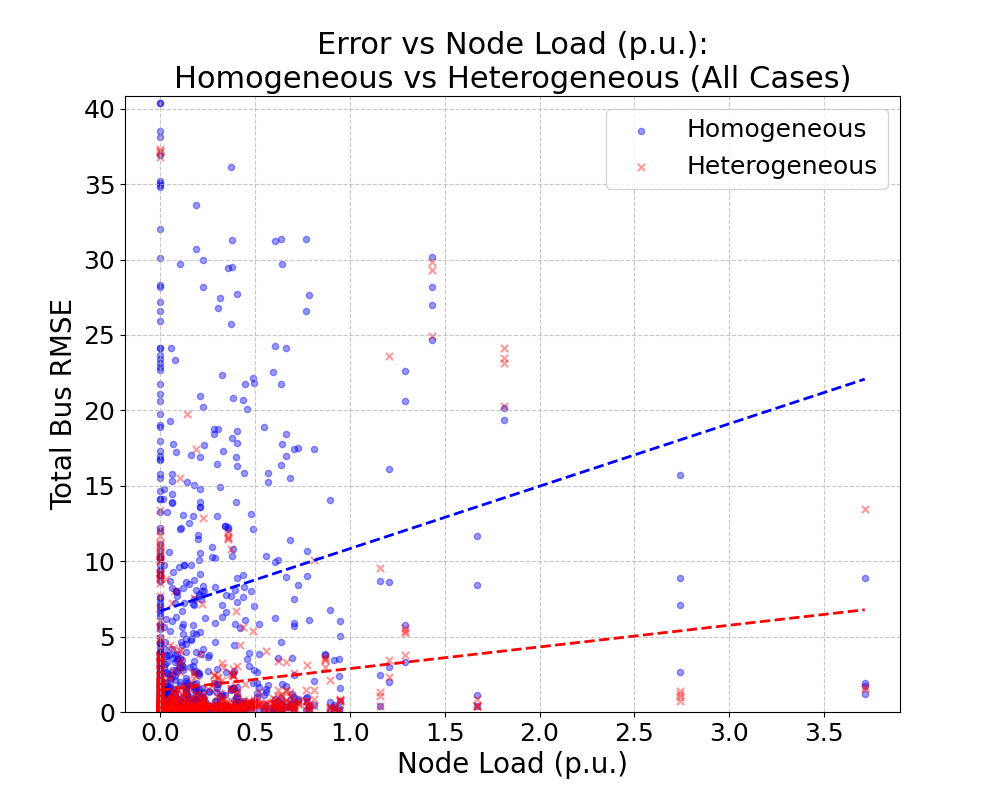}
    \caption{Model error on held-out test data across cases {14, 30, 57,118, 500, 2000} with respect to total system load across homogeneous and heterogeneous categories. Higher system load is generally correlated with error, but heterogeneous models generalize to unseen high-load cases significantly better than homogeneous ones.}
    \label{fig:homo_vs_hetero_load_generalization}
\end{figure}

\begin{figure}[!htbp]
\vspace{-0.2in}
    \centering
    \includegraphics[width=0.55\linewidth]{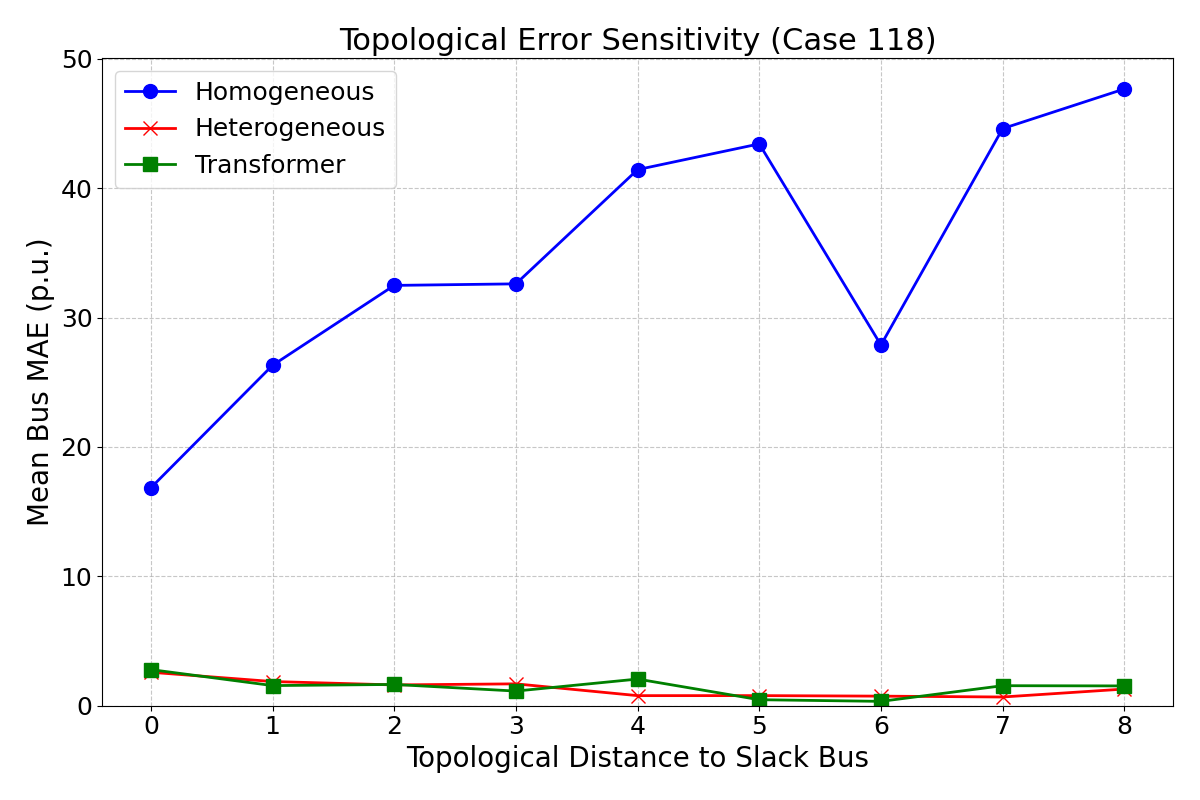}
    \caption{MAE contribution of individual buses with respect to their topological distance to reference (i.e. how many jumps there are between a given bus and a slack bus). Transformers perform significantly differently from other homogeneous models (average performance of GAT, GCN, GIN shown), with model accuracy being unaffected by distance to reference.}
    \label{fig:reference_distance}
\vspace{-0.2in}
\end{figure}


Across cases and training regimes, heterogeneous GNNs (HGT, HEAT, RGAT, HGNN) yield the most consistent accuracy–feasibility trade-offs, the most apparent under multi-topology training and topology shift. In contrast, homogeneous message-passing baselines (GCN/GAT/GIN) degrade more sharply as operating conditions shift, with noticeably weaker generalization to high load regimes (Figure \ref{fig:homo_vs_hetero_load_generalization}) and larger errors as topological distance increases (Figure \ref{fig:reference_distance}).
Among homogeneous models, the graph Transformer is the only architecture that consistently approaches heterogeneous performance, and exhibits near-zero dependence of error on topological distance, suggesting stronger global generalization; see Figure \ref{fig:reference_distance}. We provide representational evidence via intermediate-layer PCA and linear probing analyses in Appendix~\ref{app:repr_evidence}.



\begin{figure}[!htbp]
    \centering
    \begin{subfigure}[b]{\columnwidth}
        \centering
        \includegraphics[width=0.6\linewidth]{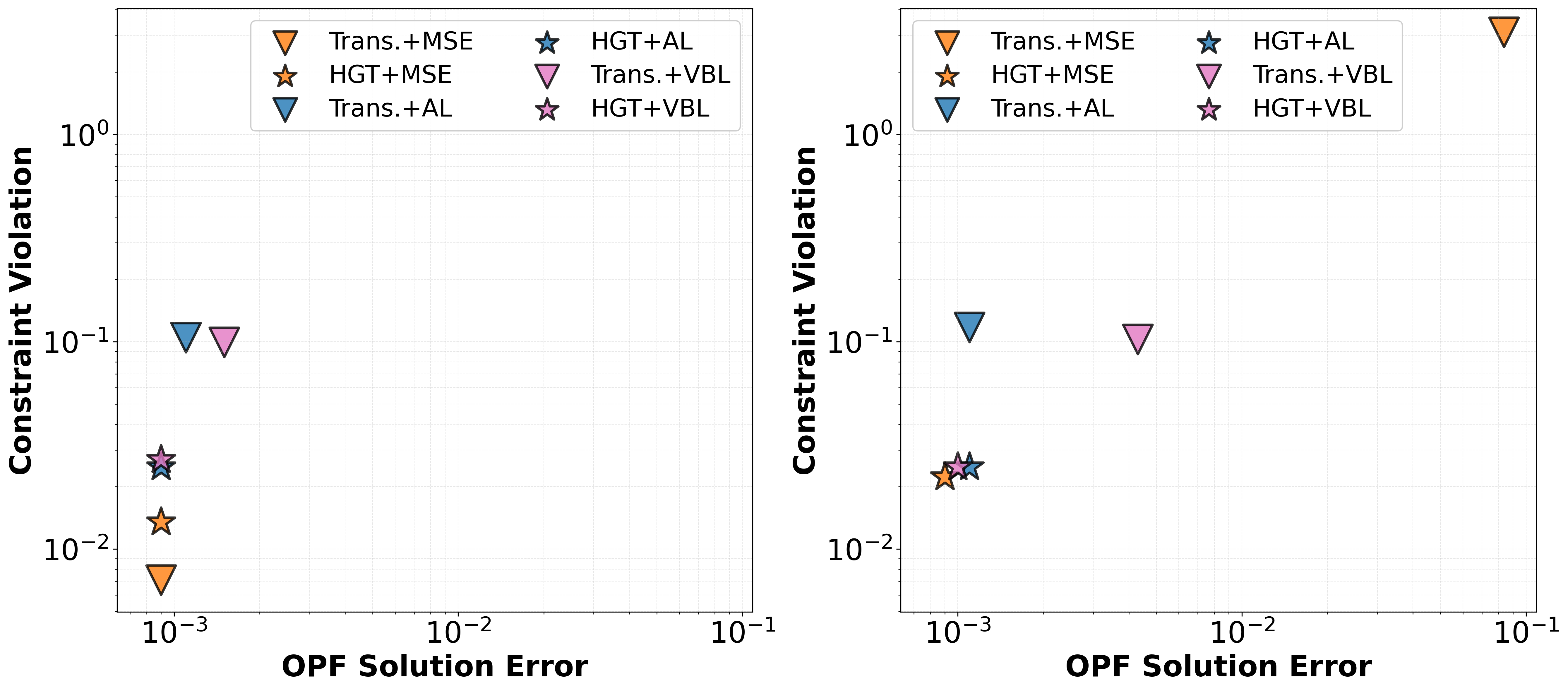}
        \caption{Evaluation on case30}
        \label{fig:loss_case30}
    \end{subfigure}
    
    \begin{subfigure}[b]{\columnwidth}
        \centering
        \includegraphics[width=0.6\linewidth]{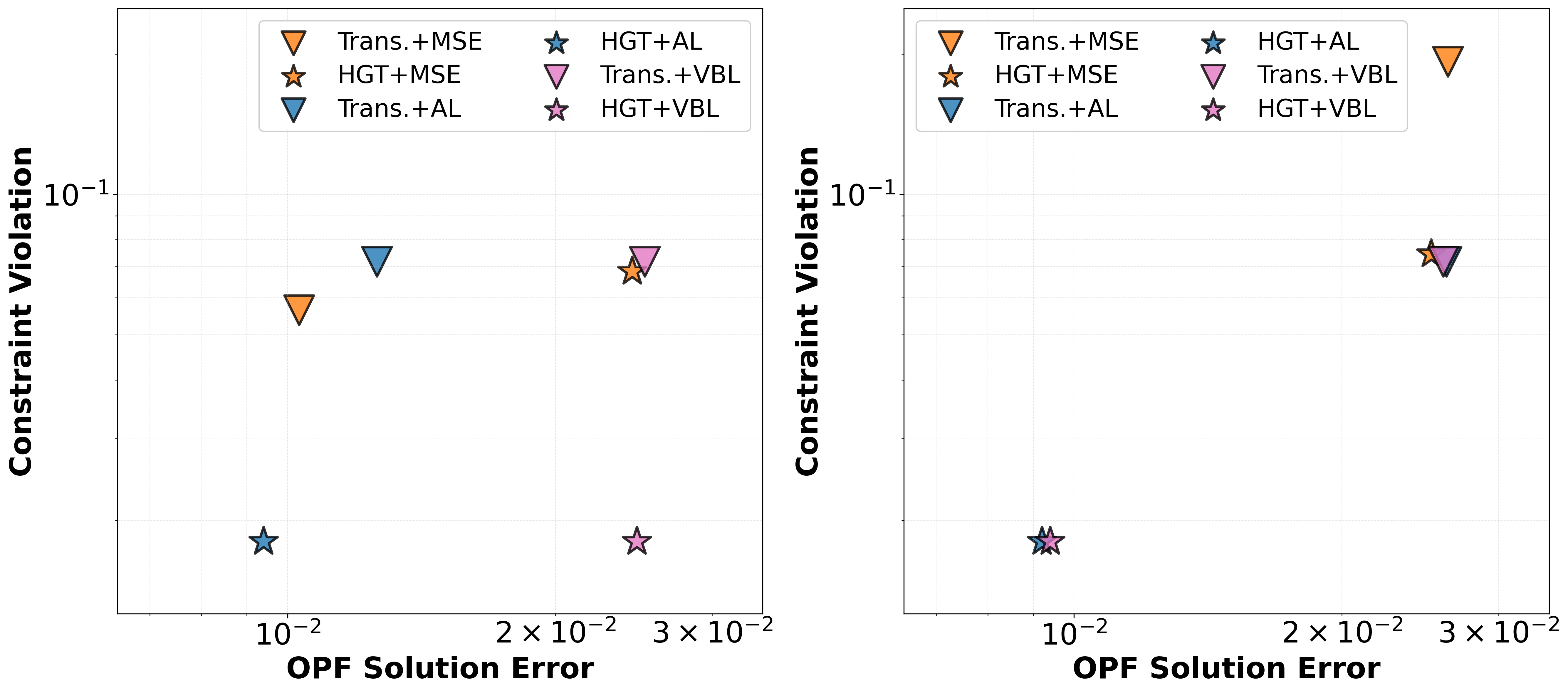}
        \caption{Evaluation on case57}
        \label{fig:loss_case57}
    \end{subfigure}
    
    \begin{subfigure}[b]{\columnwidth}
        \centering
        \includegraphics[width=0.6\linewidth]{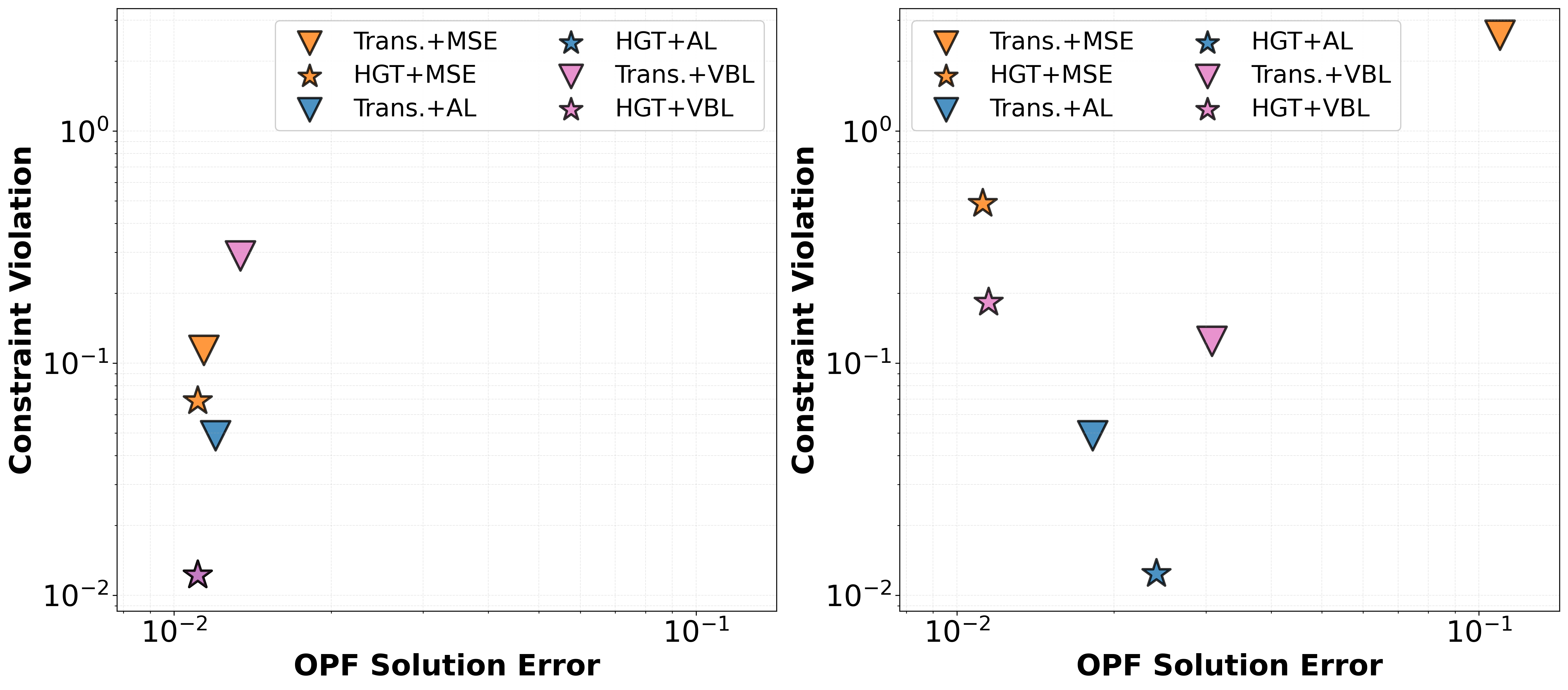}
        \caption{Evaluation on case118}
        \label{fig:loss_case118}
    \end{subfigure}
    
    \caption{Loss function comparison across architectures under equal training budgets. Left panels: single-topology training. Right panels: multi-topology training on cases 30, 57 and 118. Each architecture evaluated with three loss functions (MSE, AL, VBL).}
    \label{fig:loss_comparison}
\end{figure}

\subsection{Effect of Loss Function Design}

Figure~\ref{fig:loss_comparison} demonstrates the joint importance of architecture and loss function selection. Transformer with MSE excels on small single-topology tasks, especially on the OPF solution predictions, but degrades significantly with scale where violations increase an order of magnitude from case30 to case118 under single-topology training and struggles under multi-topology training. This implies that Transformer performs effective representation learning on label prediction, especially in small power grid topologies. HGT with AL exhibits opposite behavior. This combination shows very consistent performance across all configurations. Its superior performances are obvious under multi-topology training compared to other combinations. Notably, HGT, regardless of loss function choices, maintains lower prediction error. In general, we observe MSE produces highest violations compared to AL and VBL, while VBL offers slightly worse constraint violation controls than AL. These results demonstrate that AL loss effectively guides the training to satisfy physical constraints while minimizing the label prediction error and scales well into large-scale power grid topologies.

To evaluate the changes loss choice induces in the model's internal representation of the problem, we use linear probes \cite{alain2016understanding} to approximate a notion of how linearly classifiable different physical grid characteristics are across progressively deeper layers of a model. Two facts stand out. First, the model's representation of system characteristics becomes increasingly nonlinear with AL compared to MSE without impacting loss, pointing to the idea that AL is inducing a greater degree of physics structure in the model. The same model and layer with the use of AL loss induce markedly different structure in the activations for the same data. This additional degree of structure brought on by a more physics-aware loss may account in part for the improved generalization performance to unseen cases.

\begin{figure}[!htbp]
    \centering
    \includegraphics[width=0.5\linewidth]{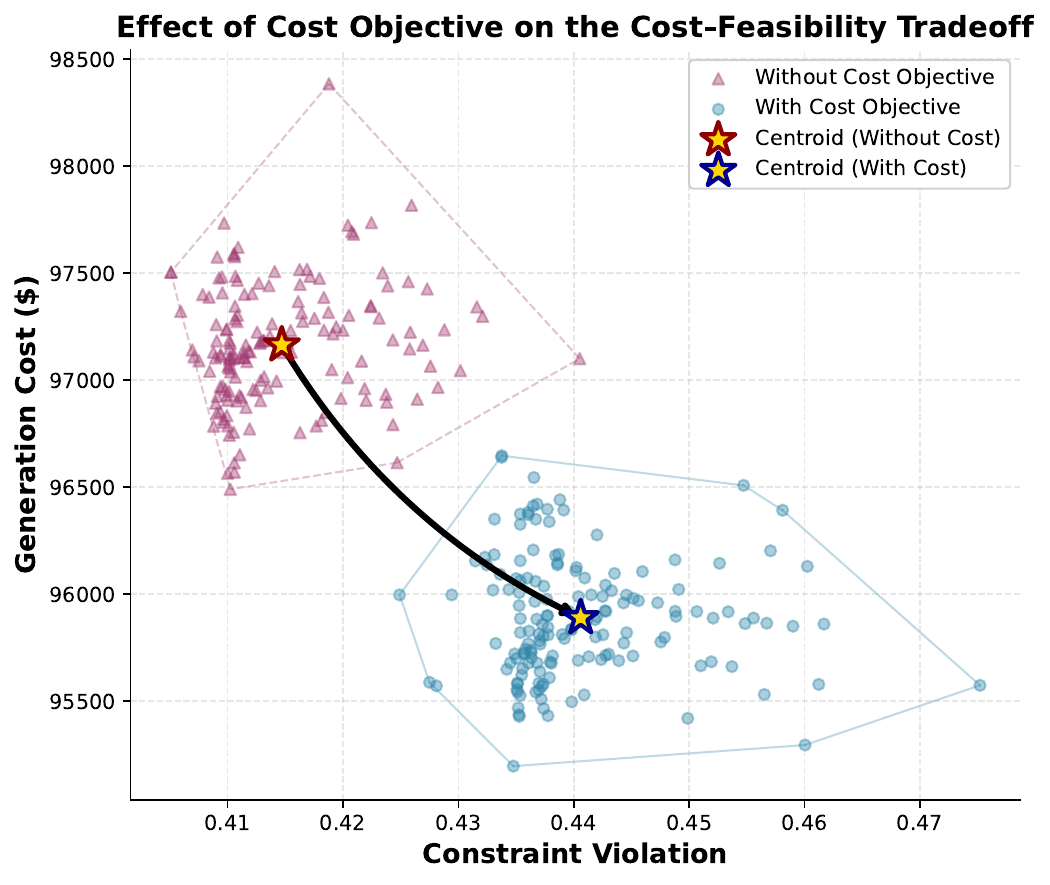}
    \caption{Cost–feasibility tradeoff with and without cost objective}
    \label{fig:cost-feasibility-tradeoff}
\vspace{-0.2in}
\end{figure}

Adding an explicit cost objective induces a systematic shift in the feasibility–optimality tradeoff rather than a uniform improvement along a single axis, as shown in Figure ~\ref{fig:cost-feasibility-tradeoff}. Without the cost term, solutions concentrate at lower violation norms while exhibiting higher and more dispersed generation costs, yet remain within a 0.1\% optimality gap of the ground-truth solution cost due to supervision on cost-optimal ACOPF solutions. Incorporating the cost objective further reduces generation cost but does so at the expense of increased constraint violation, reflected by a clear displacement of the solution centroid along the Pareto frontier in cost–violation space. This behavior indicates that, in the supervised setting, the model already internalizes cost structure through solution-level supervision, and that explicitly optimizing cost provides limited additional benefit while biasing training away from feasibility.

Additional representational diagnostics comparing AL vs.\ MSE training are provided in Appendix~\ref{app:repr_evidence}.

\subsection{Scaling Behavior Across Network Sizes}
\begin{figure}[!htbp]
    \centering
    \includegraphics[width=0.5\linewidth]{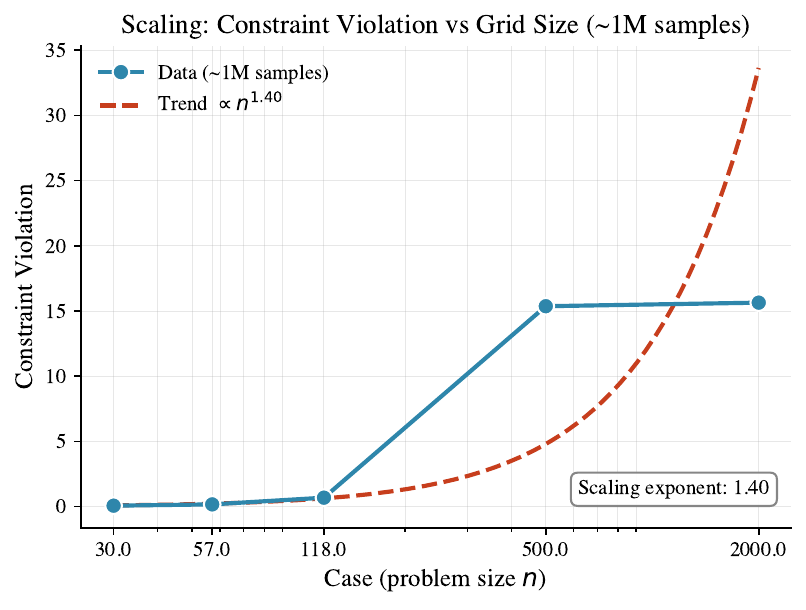}
    \caption{Topology-normalized total constraint violation across increasing power-grid sizes. Results are evaluated on a fixed budget of 1M test samples per case. The dashed curve illustrates an empirical scaling trend.}
    \label{fig:scaling}
\end{figure}
Figure~\ref{fig:scaling} shows the topology-normalized constraint violation as a function of grid size under a fixed budget of 1M test samples per topology for training Transformer with MSE loss function. Violations remain near zero for small and medium networks (case30 to case57), increasing from $\approx 0.06$ to $\approx 0.17$, indicating effective constraint control in this regime. A clear transition occurs at case118, where violations rise to $\approx 0.69$, followed by a sharp increase at case500 ($\approx 15.4$). A power-law fit across cases suggests superlinear scaling with an empirical exponent of $\approx 1.40$. However, violations saturate between case500 and case2000 ($\approx 15.6$), indicating that feasibility degradation does not increase monotonically with network size at extreme scales. The fitted exponent therefore reflects an empirical scaling trend over small-to-large systems rather than a universal law.


\begin{figure}[!htbp]
    \centering
    \includegraphics[width=0.5\linewidth]{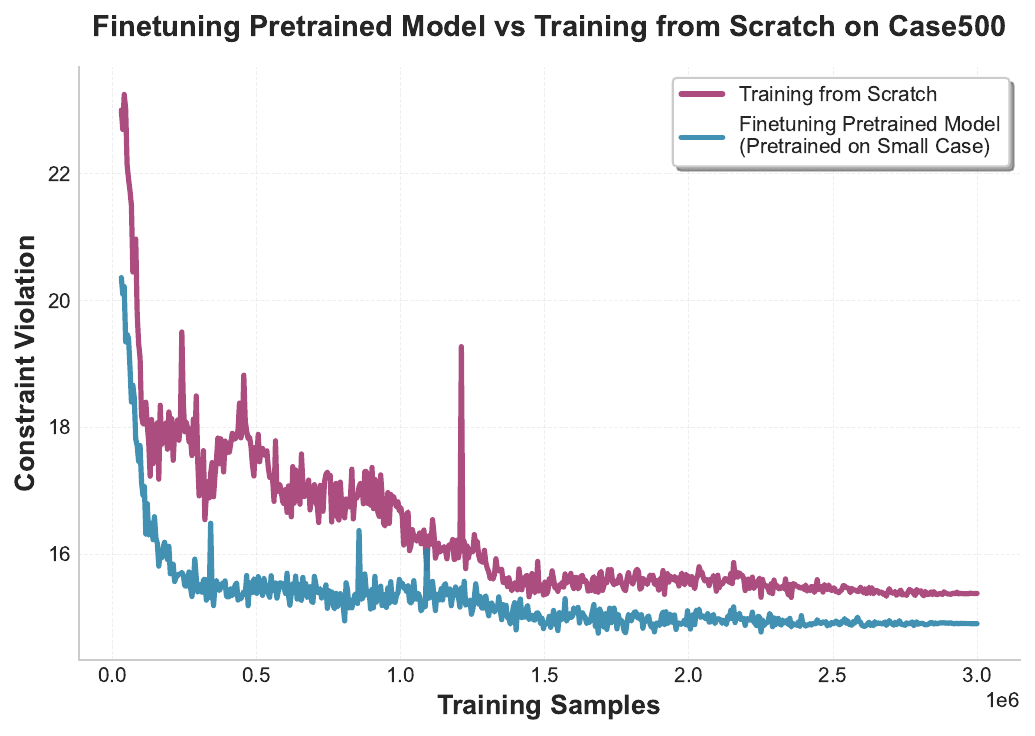}
    \caption{Constraint violation convergence on case500: fine-tuning vs. training from scratch}
    \label{fig:finetuning vs training from scratch}
\vspace{-0.2in}
\end{figure}

\subsection{Generalization Across Power System Topologies}

\textbf{Finetuning vs. Training from Scratch.}
Figure~\ref{fig:finetuning vs training from scratch} and Table~\ref{tab:convergence_speedup} jointly demonstrate that pretraining primarily improves convergence speed while also contributing to lower final violation on large topologies. On case-500, fine-tuning from a small-case pretrained model reaches the same violation threshold (i.e., 15.5) achieved by training from scratch at 1.3M samples using only 215K training steps, corresponding to an 83.6\% reduction in optimization steps. This accelerated convergence is evident in Figure~\ref{fig:finetuning vs training from scratch}, where the fine-tuned model rapidly suppresses large initial violations and stabilizes early, while training from scratch exhibits prolonged high-variance dynamics and slower decay. In addition, fine-tuning achieves a 3.1\% reduction in the final violation norm on case-500, indicating that pretraining not only reduces the sample complexity required to reach a given feasibility level but also yields improved feasibility. For the smaller case-118, fine-tuning yields both faster convergence (48.8\% fewer steps) and a larger improvement in final violation (25.1\%), suggesting that representation transfer is more effective when the target topology is closer in scale to the pretraining regime. Overall, these results indicate that foundation-style pretraining amortizes optimization effort across topologies by initializing models in a low-violation basin, yielding order-of-magnitude reductions in training steps on large grids while consistently improving final feasibility.

\begin{table}[!htbp]
\centering
\scriptsize
\setlength{\tabcolsep}{6pt}
\renewcommand{\arraystretch}{1.2}
\caption{Convergence comparison between training from scratch and fine-tuning across grid sizes. Improvements in final violation and training steps are shown as blue subscripts.}
\label{tab:convergence_speedup}
\begin{tabular}{c|cc|cc}
\toprule
\textbf{Case} 
& \multicolumn{2}{c|}{\textbf{Total Violation Norm}} 
& \multicolumn{2}{c}{\textbf{Training Steps}} \\
\cmidrule(lr){2-3}
\cmidrule(lr){4-5}
& Scratch & Finetuned 
& Scratch & Finetuned \\
\midrule
118 
& 8.986 
& $6.732_{\textcolor{blue}{(-25.1\%)}}$ 
& 3{,}000{,}000 
& $1{,}535{,}008_{\textcolor{blue}{(-48.8\%)}}$ \\
500 
& 15.387 
& $14.910_{\textcolor{blue}{(-3.1\%)}}$ 
& 1{,}310{,}048 
& $215{,}008_{\textcolor{blue}{(-83.6\%)}}$ \\
\bottomrule
\end{tabular}
\end{table}



\textbf{Zero-Shot Transfer to Unseen Topologies.} We select Transformer and HGT for zero-shot transfer evaluation as they demonstrate advantages over other architectures in prior experiments, with Transformer with MSE and HGT with AL emerging as best-in-class models. We additionally evaluate HGT with MSE to isolate the effect of architecture from loss function, given MSE's simplicity. Table~\ref{tab:cross_topology} presents the results across these configurations. Single-topology transfer (rows 1-2) demonstrates a clear solution quality-feasibility trade-off: Transformer with MSE achieves the lowest OPF solution error when transferring from case30 to larger networks but produces substantially higher constraint violations, while HGT with AL loss maintains significantly lower constraint violations at comparable solution qualities. In Multi-topology training (rows 3-5), HGT demonstrates superior transfer robustness to Transformer, achieving consistently smaller constraint violations compared to Transformer with the same MSE loss function. Comparing loss functions, AL reduces violations by an order of magnitude relative to MSE across all transfer scenarios, establishing constraint-aware training as essential for foundation model training and deployment.

\label{subsec:transfer}
\begin{table}[!htbp]
\centering
\small
\caption{Zero-shot transfer and multi-topology pretraining results. Best performance for each training→evaluation scenario shown in bold.}
\label{tab:cross_topology}
\resizebox{\columnwidth}{!}{%
\begin{tabular}{lcccccc}
\toprule
 & \multicolumn{2}{c}{\textbf{Trans.+MSE}} & \multicolumn{2}{c}{\textbf{HGT+MSE}} & \multicolumn{2}{c}{\textbf{HGT+AL}} \\
\cmidrule(lr){2-3} \cmidrule(lr){4-5} \cmidrule(lr){6-7}
\textbf{Training → Eval} & \textbf{OPF Sol. Err. } & \textbf{Viol.} & \textbf{OPF Sol. Err. } & \textbf{Viol.} & \textbf{OPF Sol. Err. } & \textbf{Viol.} \\
\midrule
case30 → case57 & \textbf{4.743} & 5.053 & 5.299 & 2.060 & 5.311 & \textbf{0.230} \\
case30 → case118 & \textbf{1.912} & 9.118 & 2.030 & 3.766 & 2.030 & \textbf{1.262} \\
\midrule
case\{57,118\} → case30 & 1.120 & 4.853 & \textbf{0.4693} & 2.286 & 0.547 & \textbf{0.239} \\
case\{30,118\} → case57 & 5.677 & 1.990 & 5.6337 & 1.4781 & \textbf{5.225} & \textbf{0.018} \\
case\{30,57\} → case118 & \textbf{2.011} & 26.295 & 6.173 & 12.26 & 2.837 & \textbf{0.907} \\
\bottomrule
\end{tabular}%
}
\vspace{-0.2in}
\end{table}

\subsection{Sensitivity to Numerical Precision}
\begin{figure}[!htbp]
    \centering
    \includegraphics[width=0.5\linewidth]{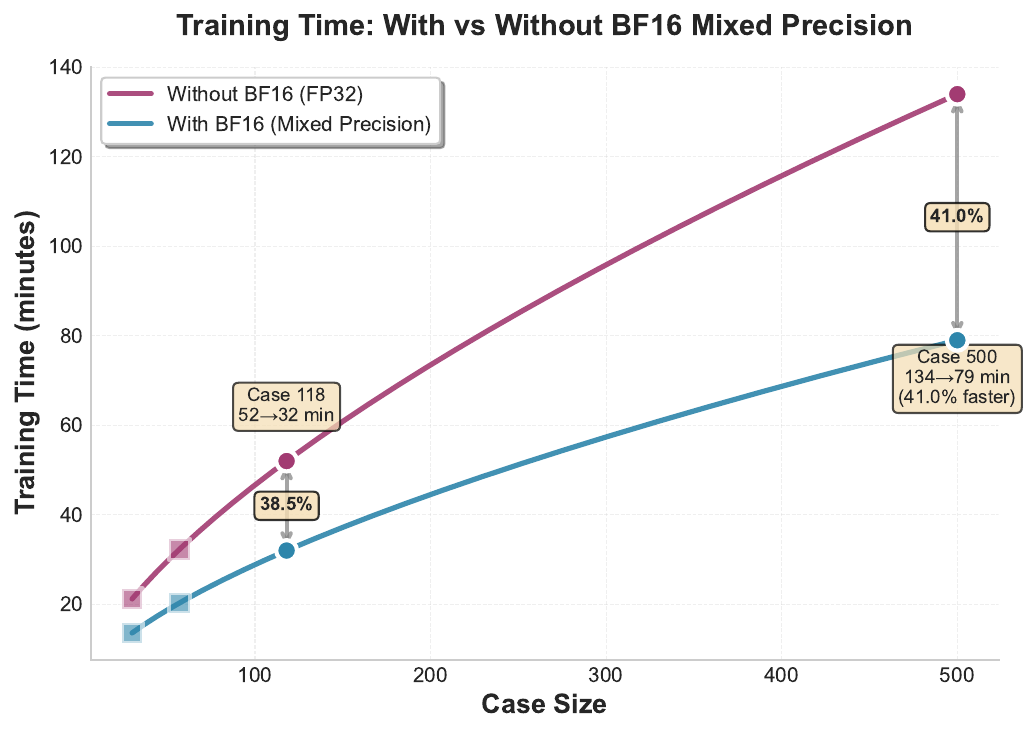}
    \caption{Training time vs. case size with and without mixed precision training(BF16)}
    \label{fig:mixed precision}
\end{figure}

Mixed-precision training with BF16 consistently accelerates optimization across network sizes, with gains increasing as problem scale grows (Figure~\ref{fig:mixed precision}). On case118, training time decreases from 52 to 32 minutes (38.5\%), while on case500 it decreases from 134 to 79 minutes (41.0\%), with absolute training timesavings rising from 20 to 55 minutes. This scaling behavior indicates that BF16 primarily reduces the dominant compute and memory costs associated with large graph message passing and constraint-aware loss evaluation, rather than providing a fixed-factor speedup. In contrast, FP32 training exhibits steeper runtime growth with case size, whereas BF16 moderates this scaling, improving effective hardware utilization on large instances. These results show that mixed precision is especially beneficial in large-grid and multi-topology training regimes central to foundation-style OPF surrogates, enabling substantial reductions in training cost without altering model structure or feasibility-aware objectives.

\subsection{Failure Modes and Diagnostics}
\label{subsec:failure}



\begin{figure}[!htbp]
    \centering
    \includegraphics[width=0.5\linewidth]{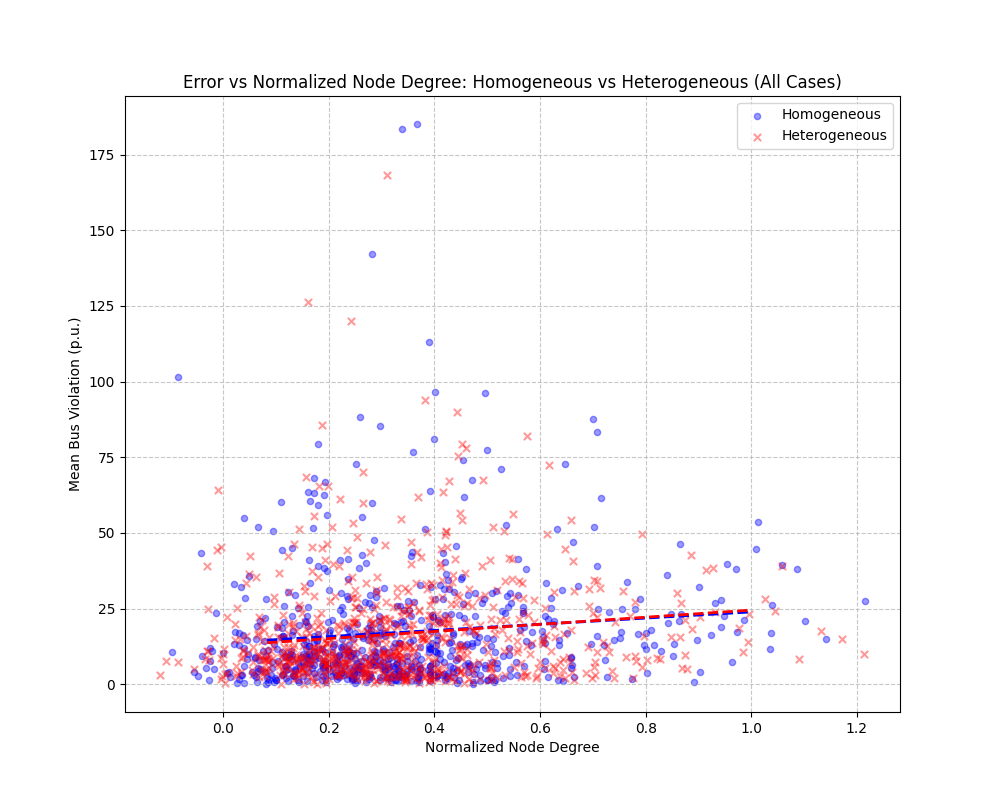}
    \vspace{-0.2in}
    \caption{Model error across categories with respect to node degree, across cases {14, 30, 57, 118, 500, 2000} with degree normalized with respect to case maximum degree. Model error by node correlates with node degree ($r=0.51$).}
    \label{fig:node_degree_vs_error}
    \vspace{-0.3in}
\end{figure}

Finally, we identify topological complexity as a crucial factor that may limit the performances of foundation-style pretraining from our analyses of failure cases. Node-level error correlates with node degree ($r=0.51$) irrespective of model type. This suggests that although we see good generalization across cases, model error tends to concentrated around topologically complex grid regions as shown in Figure~\ref{fig:node_degree_vs_error}. This diagnostic provides guidance for future dataset expansion and model improvements, and highlights the importance of combining foundation models with domain-specific safety checks in operational settings.

\section{Conclusion}
\label{sec:conclusion}

This work introduced a comprehensive benchmarking suite for evaluating surrogate model performance for AC optimal power flow, along with a systematic study comparing a range of different homogeneous and heterogeneous graph neural network architectures under three constraint-aware objectives. Our experiments yield several insights for the broader research community. 

Firstly, architecture matters and models capable of exploiting graph heterogeneity significantly outperform homogeneous variants, with HGT achieving the best overall performance. Among homogeneous models, models perform well if they are able to learn implicit encodings of node types or capture global dependencies between nodes (as in the Transformer case), making it clear that choices around the underlying graph representation have a significant impact on model training. 

Secondly, embedding constraint-awareness in the choice of loss is critical to model generalization. Using MSE as a target yields higher violations under distribution shift, with AL and VBL loss providing increasing performance improvements. Generalization performance depends primarily on the constraint-awareness of the loss and the training regime, with multi-topology pretraining having a significant impact on generalization to unseen grids particularly with an AL loss. Overall we find HGT with constraint-aware losses to be a promising approach for zero-shot transfer. 

There are some limitations to this analysis however. As is the case with many approaches to ACOPF problems, lack of data availability from real grids requires us to train on synthetic data. This increases the relevance of common benchmarks and benchmarking tools to the broader research community. However, even though several of the topologies selected are standard IEEE test systems and others are generated using real grid statistics\cite{birchfield2016grid}, there may be physical grid conditions not captured by the synthetic data. This may limit generalization to real-world scenarios without fine-tuning, although we demonstrate that multi-topology pretraining may effectively limit the amount of fine-tuning data required.

As grid modernization continues to increase operational complexity, fast and reliable ACOPF surrogates will be critical for operators to strengthen the resilience of the grid, and this suite contributes a standardized set of research benchmarks and models in support of that aim.

\section*{Acknowledgments}
This material is based upon work supported by Laboratory Directed Research and Development (LDRD) funding from Argonne National Laboratory, provided by the Director, Office of Science, of the U.S. Department of Energy under contract DE-AC02-06CH11357.

An award of computer time was provided by the ASCR Leadership Computing Challenge (ALCC) program. This research used resources of the Argonne Leadership Computing Facility, which is a U.S. Department of Energy Office of Science User Facility operated under contract DE-AC02-06CH11357.

This research used resources of the National Energy Research Scientific Computing Center (NERSC), a Department of Energy User Facility using NERSC award ALCC-ERCAP0038201.

\bibliographystyle{plain}
\bibliography{references}

\begin{mdframed}
The submitted manuscript has been created by UChicago Argonne, LLC, Operator of Argonne National Laboratory (``Argonne”). Argonne, a U.S. Department of Energy Office of Science laboratory, is operated under Contract No. DE-AC02-06CH11357. The U.S. Government retains for itself, and others acting on its behalf, a paid-up nonexclusive, irrevocable worldwide license in said article to reproduce, prepare derivative works, distribute copies to the public, and perform publicly and display publicly, by or on behalf of the Government. The Department of Energy will provide public access to these results of federally sponsored research in accordance with the DOE Public Access Plan (http://energy.gov/downloads/doe-public-access-plan).
\end{mdframed}

\appendix

\section{AC Branch-flow Formulation}
The active and reactive power flows (from $i$ to $j$) are defined as
\begin{align}
    P_{ij}(V,\theta) &:= V_i^2 G_{ij}  - V_i V_j (G_{ij}\cos(\theta_{ij}) + B_{ij} \sin(\theta_{ij})), \\
    Q_{ij}(V,\theta) &:= -V_i^2 B_{ij}  - V_i V_j (G_{ij} \sin(\theta_{ij}) - B_{ij}\cos(\theta_{ij})),\label{eq:pf3}
\end{align}
where $P_{d,i}$, $Q_{d,i}$ are the active and reactive loads for bus $i \in \mathcal{N}$, $V_i$, $\theta_i$ are the voltage magnitude and angle, $\theta_{ij} = \theta_i - \theta_j$, and $Y_{ij} = G_{ij} + jB_{ij}$ is the element of the bus admittance matrix.

\section{Full Node/Edge Features in Graph Representation}
\label{s:full_feature}

Nodes and edges correspond to physical components of the grid. The node set $\mathcal{V}$ comprises four node types:
\begin{itemize}
    \item \emph{Buses}: voltage bounds $(V_i^{min}, V_i^{max})$ and a one-hot bus type (PQ/PV/slack).
    \item \emph{Generators}: $(P_g^{min}, P_g^{max}, Q_g^{min}, Q_g^{max})$ and cost coefficients.
    \item \emph{Loads}: $(P_{d,i}, Q_{d,i})$.
    \item \emph{Shunts}: $(G_s, B_s)$.
\end{itemize}
Edges $\mathcal{E}$ correspond to transmission lines and transformers, representing connectivity between buses. Edge features include line parameters derived from the admittance matrix $Y_{ij}$ and thermal limits $S_{ij}^\text{max}$.

\section{Baseline Model Update Equations}
\label{app:baseline_architectures}

\subsection{Homogeneous Models.} 
Homogeneous GNNs use a shared set of parameters across all nodes and edges, and do not explicitly condition their message functions on node or edge types.

\textbf{GCN:} 
Graph convolutional networks~\cite{kipf2016semi} perform normalized neighborhood aggregation. Using self-loops and $\tilde{A}=A+I$, $\tilde{D}_{ii}=\sum_j \tilde{A}_{ij}$, the layer update can be written as 
\begin{equation*}
    h_i^{\ell + 1} = \sigma \left(W^{\ell} \sum_{i,j \in N(i)} \frac{1}{\sqrt{d_i d_j}} h^{\ell}_j \right).
\end{equation*}
Note that GCNs only capture adjacency, ignoring edge features entirely.

\textbf{GAT:} Graph attention networks~\cite{velivckovic2017graph} learn attention weights over neighbors. For a single head, 
\begin{equation*}
    \alpha^{\ell}_{ij} = \mathrm{softmax}(\mathrm{LeakyReLU}(a[Wh^{\ell}_i || Wh^{\ell}_j])),  
\end{equation*}
with layer updates given by:
\begin{equation*}
    h^{\ell + 1}_i = \sigma(\sum_{j\in \mathcal{N}} \alpha^{\ell}_{ij}Wh^{\ell}_j ).
\end{equation*}
 
\textbf{GIN:} Graph Isomorphism networks\cite{xu2018powerful} use an MLP after aggregating node features over a neighborhood like:
\begin{equation*}
    h^{\ell}_i = \mathrm{MLP}^{\ell}((1 + \epsilon^{\ell})h^{\ell}_i + \sum_{j\in \mathcal{N}(i)} h^{\ell}_j ).
\end{equation*} 

\textbf{Graph Transformer:} 
Graph transformers~\cite{yun2019graph} do not explicitly encode node or edge type information, but do have global attention, which may give a model a way to capture global node-node relationships that would otherwise be captured explicitly by node type. As in \cite{yun2019graph} we model the query/key/value tuple for a given layer like: 
\begin{equation*} 
    Q^{(\ell)} = H^{(\ell)}W_Q^{(\ell)},\quad K^{(\ell)} = H^{(\ell)}W_K^{(\ell)},\quad V^{(\ell)} = H^{(\ell)}W_V^{(\ell)}, 
\end{equation*} 
where $H^{(\ell)}\in\mathbb{R}^{|\mathcal{V}|\times d_\ell}$ stacks node embeddings.
Attention is then computed as 
\begin{equation*} 
    \mathrm{Attn}(Q,K,V)=\mathrm{softmax}\!\left(\frac{QK^\top}{\sqrt{d_k}} + M\right)V, 
\end{equation*} 
where $M$ is an attention mask encoding adjacency-restricted attention.

\subsection{Heterogeneous Models.} 
Fully heterogeneous GNNs encode both node types $t_i$ and edge types $r_{ij}$, using type-specific transformations. This is well-matched to power-grid structure, where buses, generators, loads, and shunts play distinct physical roles and participate in different subsets of constraints. 

\textbf{RGAT:} Relational graph attention network \cite{busbridge2019relationalgraphattentionnetworks} attempt to explicitly encode edge relations of different types by giving each edge category its own projection matrices
\begin{equation*}
    m^{\ell}_{ij} = W^{\ell}_{r_{ij}} h^{\ell}_j, \;
    \alpha^{\ell}_{ij} = softmax(score(h^{\ell}_i, h^{\ell}_j, r(i,j), e_{ij}))
\end{equation*}
where layer updates are then given by
\begin{equation*}
    h^{\ell}_i = \sigma(\sum_{j\in \mathcal{N}}(\alpha^{\ell}_{ij} m^{\ell}_{ij})).
\end{equation*}

\textbf{HeteroGNN:} Heterogeneous graph neural networks \cite{zhang2019hetgnn}, are a straightforward extension of GNNs that learn a feature projection for each node type and aggregate messages from node-type specific neighborhoods. Layer updates are given by:
\begin{equation*}
    h^{\ell + 1}_i = \sigma (W^{\ell}_t h^{\ell}_i + \sum_{\tau \in \mathcal{T} }\mathrm{AGG}(h^{\ell}_\tau )
\end{equation*}
where $\tau \in \mathcal{T}$ is a typed node from a neighborhood of adjacent nodes of that type.

\textbf{HGT:} Heterogeneous Graph Transformer (HGT)~\cite{hu2020heterogeneous} combines typed message passing with multi-head attention. For node $i$ attending to neighbor $j$ along edge relation $r_{ij}$ with node types $t_i,t_j$, HGT uses type-dependent projections giving Q/K/V like:
\begin{equation*}
    q_i^{(\ell)} = W_Q^{(\ell,t_i)} h_i^{(\ell)}, k_j^{(\ell)} = W_K^{(\ell,t_j)} h_j^{(\ell)}, v_j^{(\ell)} = W_V^{(\ell,t_j)} h_j^{(\ell)}
\end{equation*}
An attention model that attends to type-specific connections like:
\begin{equation*} 
    s_{ij}^{\ell} = \frac{\left(q_i^{(\ell)}\right)^\top \left(W_A^{(\ell,r_{ij})} k_j^{\ell}\right)}{\sqrt{d_k}}, \;
    \alpha_{ij}^{(\ell)} = \mathrm{softmax}_{j \in N(i)}(s_{ij}^{\ell}).
\end{equation*}

\textbf{HEAT:} Heterogeneous edge attention transformers ~\cite{mo2022heat} are a category of transformer that augments typed attention with additional edge-enhancement. HEAT follows the same typed message passing template as above, with attention scores and messages explicitly conditioned on $(t_i,t_j,r_{ij},e_{ij})$: 
\begin{equation*} 
    m_{ij}^{(\ell)} = \phi^{(\ell)}\!\left(h_i^{(\ell)},h_j^{(\ell)},e_{ij},t_i,t_j,r_{ij}\right),\; h_i^{(\ell+1)}=\sigma\!\left(\sum_{j \in N(i)} \alpha_{ij}^{(\ell)} m_{ij}^{(\ell)}\right), 
\end{equation*} 
where $\phi^{(\ell)}(\cdot)$ is a learnable message function and $\alpha_{ij}^{(\ell)}$ is computed by a typed attention mechanism.

\section{Hyperparameter Configurations}
\label{app:hyperparams}
\subsection{Hyperparameter optimization settings.} 
Tables~\ref{tab:hpo_stage1} and~\ref{tab:hpo_stage2} detail the hyperparameter ranges and sampling strategies for Stage 1 (architecture and learning sweep) and Stage 2 (loss function sweep), respectively.

\begin{table}[h]
\centering
\small
\caption{Stage 1 Hyperparameter Ranges for Architecture Sweep (50 configurations per architecture)}
\label{tab:hpo_stage1}
\begin{tabular}{p{0.25\linewidth}p{0.70\linewidth}}
\toprule
\textbf{Architecture} & \textbf{Hyperparameters and Ranges} \\
\midrule
GAT, GCN, GIN, Transformer, HGNN & 
\begin{tabular}[t]{@{}l@{}}
Number of layers: $[3, 4, 6]$ \\
Hidden dimensions: $[128, 256, 512]$\\
Dropout: $[0.0, 0.3]$, uniform \\
Learning rate: $[1\times10^{-4}, 3\times10^{-3}]$, log-uniform \\
Weight decay: $[1\times10^{-6}, 1\times10^{-2}]$, log-uniform \\
Gradient clip value: $[0.5, 2.0]$, uniform \\
\end{tabular} \\
\midrule
HEAT& 
\begin{tabular}[t]{@{}l@{}}
Number of layers: $[3, 4, 6]$ \\
Hidden dimensions: $[128, 256, 512]$\\
Dropout: $[0.0, 0.3]$, uniform \\
Number of attention heads: $[2, 4, 8]$ \\
Learning rate: $[1\times10^{-4}, 3\times10^{-3}]$, log-uniform \\
Weight decay: $[1\times10^{-6}, 1\times10^{-2}]$, log-uniform \\
Gradient clip value: $[0.5, 2.0]$, uniform \\
\end{tabular} \\
\midrule
HGT& 
\begin{tabular}[t]{@{}l@{}}
Number of layers: $[3, 4, 6]$ \\
Hidden dimensions: $[128, 256, 512]$\\
Dropout: $[0.0, 0.3]$, uniform \\
Number of heads: $[2, 4, 8]$ \\
Learning rate: $[1\times10^{-4}, 3\times10^{-3}]$, log-uniform \\
Weight decay: $[1\times10^{-6}, 1\times10^{-2}]$, log-uniform \\
Gradient clip value: $[0.5, 2.0]$, uniform \\
\end{tabular} \\
\midrule
RGAT& 
\begin{tabular}[t]{@{}l@{}}
Number of layers: $[3, 4, 6]$ \\
Hidden dimensions: $[128, 256, 512]$\\
Dropout: $[0.0, 0.3]$, uniform \\
Number of heads: $[2, 4, 8]$ \\
Edge dim: $[2, 4, 8]$ \\
Learning rate: $[1\times10^{-4}, 3\times10^{-3}]$, log-uniform \\
Weight decay: $[1\times10^{-6}, 1\times10^{-2}]$, log-uniform \\
Gradient clip value: $[0.5, 2.0]$, uniform \\
\end{tabular} \\
\bottomrule
\end{tabular}
\end{table}

\begin{table}[h]
\centering
\small
\caption{Stage 2 Hyperparameter Ranges for Loss Function Sweep (40 configurations per loss function)}
\label{tab:hpo_stage2}
\begin{tabular}{p{0.25\linewidth}p{0.70\linewidth}}
\toprule
\textbf{Loss Function} & \textbf{Hyperparameters and Ranges} \\
\midrule
Augmented Lagrangian (AL), Violation-Based Lagrangian (VBL) & 
\begin{tabular}[t]{@{}l@{}}
Step size for Lagrangian multipliers update: \\
$[1\times10^{-5}, 1\times10^{-1}]$, uniform\\
Exponential moving average factor: $[0.0, 1.0]$: uniform \\
Warm-up samples: \\ $[250K, 500K, 750K, 1M, 1.25M, 1.5M]$ \\
Penalty check samples:\\ $[250K, 500K, 750K, 1M, 1.25M, 1.5M]$ \\
Multiplier check samples:\\ $[250K, 500K, 750K, 1M, 1.25M, 1.5M]$ \\
\end{tabular} \\
\bottomrule
\end{tabular}
\end{table}

\subsection{Hyperparameter sensitivity and robustness.}
\begin{figure}[!htbp]
    \centering
    \begin{subfigure}[b]{0.6\linewidth}
        \centering
        \includegraphics[width=\linewidth]{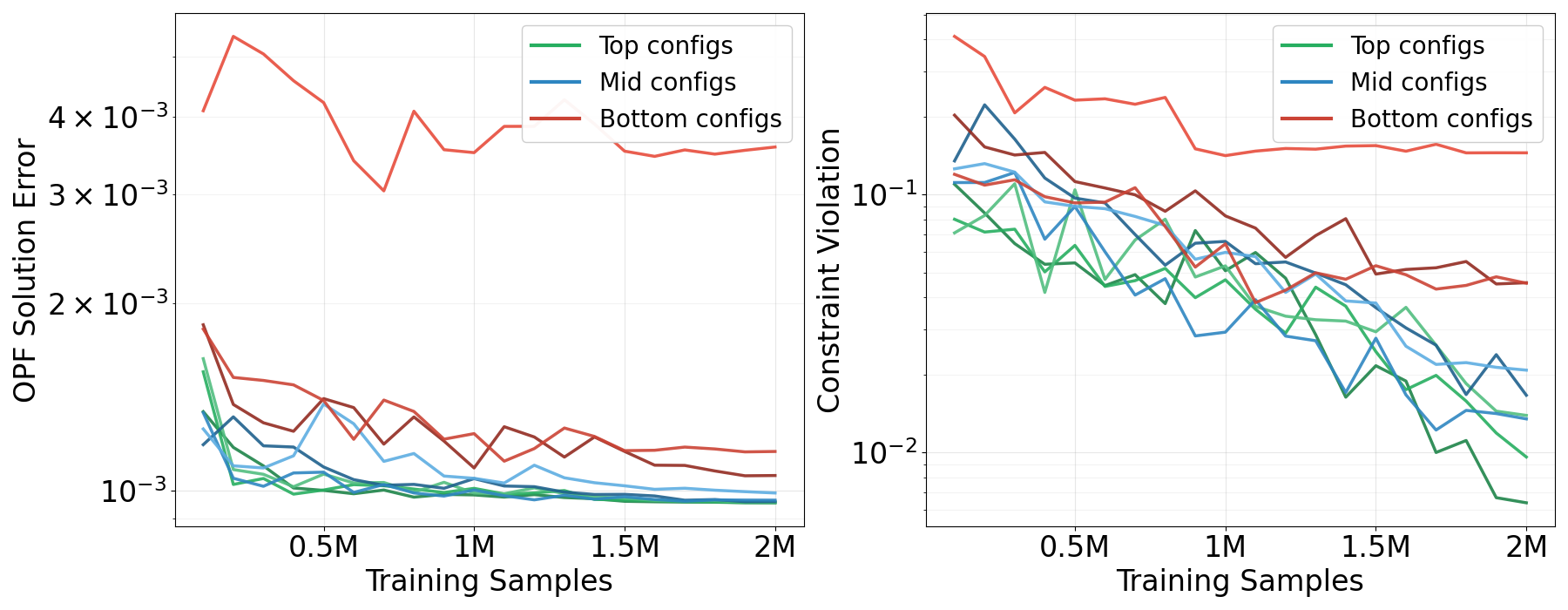}
        \caption{Transformer with MSE loss}
        \label{fig:hpo_transformer_mse}
    \end{subfigure}
    
    \begin{subfigure}[b]{0.6\linewidth}
        \centering
        \includegraphics[width=\linewidth]{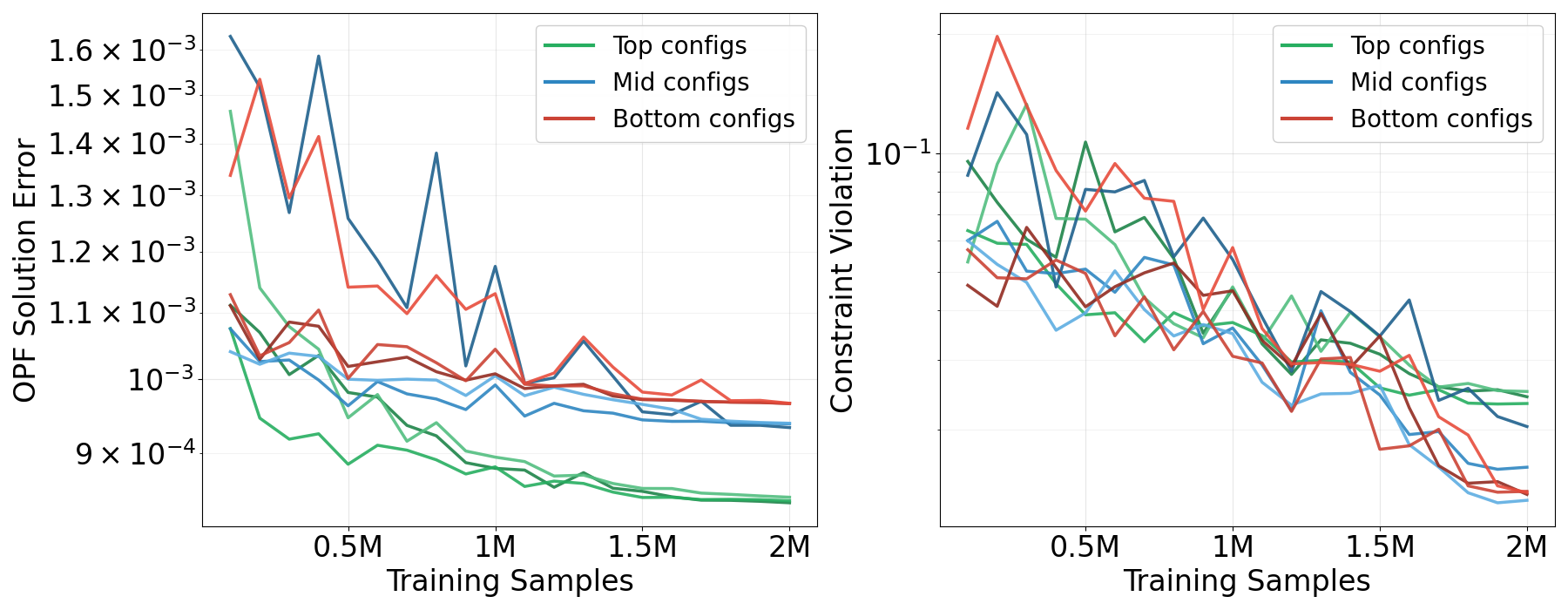}
        \caption{HGT with MSE loss}
        \label{fig:hpo_hgt_mse}
    \end{subfigure}
    
    \begin{subfigure}[b]{0.6\linewidth}
        \centering
        \includegraphics[width=\linewidth]{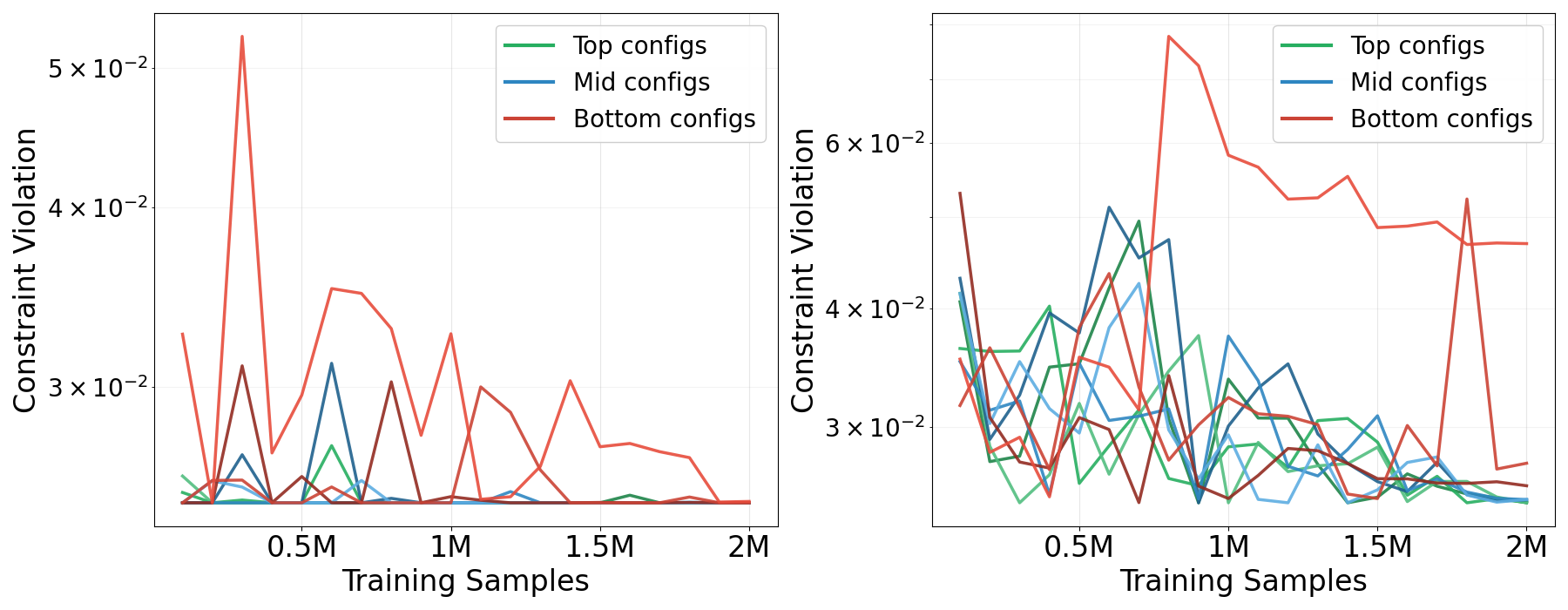}
        \caption{HGT with AL (left) and VBL (right) loss functions}
        \label{fig:hpo_hgt_al_vbl}
    \end{subfigure}
    
    \caption{Hyperparameter optimization sensitivity analysis across 2M training samples. (a) Transformer with MSE loss exhibits high performance spread in OPF solution error with rapid early convergence. (b) HGT with MSE loss demonstrates tighter performance clustering and superior constraint satisfaction. (c) HGT loss function comparison reveals high sensitivity to hyperparameters in both AL and VBL methods, with comparable best-case performance.}
    \label{fig:hpo_comprehensive}
\end{figure}

Figure~\ref{fig:hpo_comprehensive} show training progression for nine representative configurations grouped by performance selected from stable runs. The first two plots compare stage 1 sweep with Transformer and HGT trained with MSE loss function. The Transformer architecture converges rapidly within the first 500K samples, but exhibits a wide performance spread in OPF solution error. The constraint violations decrease more gradually with best configurations. HGT shows tighter clustering compared to Transformer and achieves superior constraint satisfaction in top configurations. In stage 2 sweep, we compare AL and VBL loss functions on HGT. Both exhibit high sensitivity to hyperparameters with substantial performance variance. AL's best configuration achieves consistently better constraint violation compared to worse configurations. VBL, however, shows a much noisier convergence. The performance clustering differences between architectures and high loss function sensitivity highlight the critical importance of comprehensive HPO for constrained optimization problems.

\section{Additional Results}
\label{app:additional_experiments}

\subsection{Single- and Multi-Topology Baselines.} Tables~\ref{tab:case30_comparison}-\ref{tab:case118_comparison} compare single-topology and multi-topology pretraining performance across all eight architectures using MSE loss. Each table reports OPF solution errors (OPF Sol. Err. ), constraint violations (Viol.), and cost difference (Cost Diff.) for held-out test samples from the evaluated topology. Cost difference quantifies the actual and percentage deviation between generation costs computed from predicted power generation and ground truth optimal power generation, providing an economic measure of solution quality.

\begin{table}[!htbp]
\centering
\scriptsize
\caption{Evaluation on case30 from single- and multi-topology training with MSE loss function.}
\label{tab:case30_comparison}
\resizebox{\columnwidth}{!}{%
\begin{tabular}{lccc|ccc}
\toprule
\textbf{Trained on} & \multicolumn{3}{c}{\textbf{Single (case30)}} & \multicolumn{3}{c}{\textbf{Multi (30+57+118)}} \\
\cmidrule(lr){2-4} \cmidrule(lr){5-7}
\textbf{Arch.} & \textbf{OPF Sol. Err. } & \textbf{Viol.} & \textbf{Cost Diff.} & \textbf{OPF Sol. Err. } & \textbf{Viol.} & \textbf{Cost Diff.} \\
\midrule
GCN & 0.0876 & 2.231 & -583.65(-35.91\%) & 0.0934 & 2.239 & -591.29(-36.38\%) \\
GAT & 0.0849 & 2.790 & -571.59(-35.17\%) & 0.1023 & 2.3976 & -506.08(-31.14\%) \\
GIN & 0.0801 & 2.885 & -371.65(-22.87\%) & 0.2936 & 3.262 & 1029.97(63.37\%) \\
Trans. & 0.0009 & 0.0071 & -0.08(-0.01\%) & 0.0836 & 3.114 & -184.36(-11.34\%) \\
\midrule
RGAT & 0.0009 & 0.0209 & -3.14(-0.19\%) & 0.0116 & 0.21 & 415.07(25.5\%) \\
HEAT & 0.0008 & 0.0214 & -3.08(-0.19\%) & 0.0009 & 0.0482 & -6.48(-0.40\%) \\
HGT & 0.0009 & 0.0135 & -3.19(-0.2\%) & 0.0009 & 0.0223 & -4.34(-0.27\%) \\
HGNN & 0.0011 & 0.0303 & -47.21(-2.9\%) & 0.0037 & 0.1303 & 193.27(11.87\%) \\
\bottomrule
\end{tabular}%
}
\end{table}

\begin{table}[!htbp]
\centering
\scriptsize
\caption{Evaluation on case57 from single- and multi-topology training with MSE loss function.}
\label{tab:case57_comparison}
\resizebox{\columnwidth}{!}{%
\begin{tabular}{lccc|ccc}
\toprule
\textbf{Trained on} & \multicolumn{3}{c}{\textbf{Single (case57)}} & \multicolumn{3}{c}{\textbf{Multi (30+57+118)}} \\
\cmidrule(lr){2-4} \cmidrule(lr){5-7}
\textbf{Arch.} & \textbf{OPF Sol. Err. } & \textbf{Viol.} & \textbf{Cost Diff.} & \textbf{OPF Sol. Err. } & \textbf{Viol.} & \textbf{Cost Diff.} \\
\midrule
GCN & 0.0505 & 1.149 & -4441.49(-9.13\%) & 0.200 & 2.952 & -11621.11(-23.89\%) \\
GAT & 0.1263 & 2.620 & -7526.33(-15.47\%) & 0.0143 & 0.5341 & 274.15(0.56\%) \\
GIN & 0.0397 & 2.438 & -108.03(-0.22\%) & 0.3018 & 1.4830 & 7438.85(15.29\%) \\
Trans. & 0.0103 & 0.0565 & -48.16(-0.10\%) & 0.0263 & 0.1926 & 498.63(1.03\%) \\
\midrule
RGAT & 0.0085 & 0.0662 & -48.76(-0.10\%) & 0.0215 & 0.3953 & 3923.45(8.04\%) \\
HEAT & 0.0085 & 0.0596 & -44.68(-0.09\%) & 0.0085 & 0.0855 & -31.5(-0.06\%) \\
HGT & 0.0244 & 0.0684 & -136.02(-0.28\%) & 0.0252 & 0.0745 & -192.29(-0.39\%) \\
HGNN & 0.0271 & 0.0710 & -688.55(-1.41\%) & 0.0480 & 1.0256 & -2934.83(-6.02\%) \\
\bottomrule
\end{tabular}%
}
\end{table}

\begin{table}[!htbp]
\centering
\scriptsize
\caption{Evaluation on case118 from single- and multi-topology training with MSE loss function.}
\label{tab:case118_comparison}
\resizebox{\columnwidth}{!}{%
\begin{tabular}{lccc|ccc}
\toprule
\textbf{Trained on} & \multicolumn{3}{c}{\textbf{Single (case118)}} & \multicolumn{3}{c}{\textbf{Multi (30+57+118)}} \\
\cmidrule(lr){2-4} \cmidrule(lr){5-7}
\textbf{Arch.} & \textbf{OPF Sol. Err. } & \textbf{Viol.} & \textbf{Cost Diff.} & \textbf{OPF Sol. Err. } & \textbf{Viol.} & \textbf{Cost Diff.} \\
\midrule
GCN & 0.111 & 8.434 & -17806.35(-11.64\%) & 0.2315 & 19.308 & -2429.87(-1.59\%) \\
GAT & 0.243 & 22.34 & -40301.82(-26.33\%) & 1.477 & 65.92 & -64212.43(-41.96\%) \\
GIN & 0.0925 & 4.013 & -7844.08(-5.13\%) & 0.1197 & 3.9951 & -4695.77(-3.07\%) \\
Trans. & 0.0114 & 0.1136 & -224.07(-0.15\%) & 0.1097 & 2.5862 & -16741.81(-10.94\%) \\
\midrule
RGAT & 0.0102 & 0.8198 & -300.40(-0.20\%) & 0.0440 & 1.121 & -13787.74(-9.00\%) \\
HEAT & 0.0101 & 0.2278 & -292.31(-0.19\%) & 0.0116 & 0.3140 & -4240.26(-2.77\%) \\
HGT & 0.0111 & 0.0687 & -297.97(-0.19\%) & 0.0112 & 0.4872 & 334.86(0.22\%) \\
HGNN & 0.0141 & 0.1718 & -2251.08(-1.47\%) & 0.1070 & 2.1519 & -30316.36(-19.8\%) \\
\bottomrule
\end{tabular}%
}
\end{table}

\subsection{Effect of Loss Function Design.} Tables~\ref{tab:constraint_violations_case30}-\ref{tab:constraint_violations_case118} compare loss function effects on selected architectures (Transformer and HGT) across three loss functions (MSE, AL, VBL). The same metrics as Tables~\ref{tab:case30_comparison}-\ref{tab:case118_comparison} are reported. Models are evaluated on case30, case57, and case118 test sets respectively, with left columns showing single-topology training results and right columns showing multi-topology training (cases 30, 57, and 118) results. Cost difference represents the actual and percentage deviation between generation costs computed from predicted power generation and ground truth optimal power generation.

\begin{table}[!htbp]
\centering
\scriptsize
\caption{Evaluation on case30 from single- and multi-topology training with selective models and three different loss functions.}
\label{tab:constraint_violations_case30}
\resizebox{\columnwidth}{!}{%
\begin{tabular}{llccc|ccc}
\toprule
\multicolumn{2}{c}{\textbf{Trained on}} & \multicolumn{3}{c}{\textbf{Single (case30)}} & \multicolumn{3}{c}{\textbf{Multi (30+57+118)}} \\
\cmidrule(lr){3-5} \cmidrule(lr){6-8}
\textbf{Arch.} & \textbf{Loss} & \textbf{OPF Sol. Err. } & \textbf{Viol.} & \textbf{Cost Diff.} & \textbf{OPF Sol. Err. } & \textbf{Viol.} & \textbf{Cost Diff.} \\
\midrule
\multirow{3}{*}{GAT} 
& MSE & 0.0849 & 2.790 & -571.59(-35.17\%) & 0.1023 & 2.3976 & -506.08(-31.14\%) \\
& AL & 0.1122 & 0.0995 & -394.35(-24.26\%) & 0.1090 & 0.1177 & -409.07(-25.17\%) \\
& VBL & 0.0867 & 1.9376 & -621.35(-38.23\%) & 0.4245 & 3.039 & 247(15.20\%) \\
\midrule
\multirow{3}{*}{Trans.} 
& MSE & 0.0009 & 0.0071 & -0.08(-0.01\%) & 0.0836 & 3.114 & -184.36(-11.34\%) \\
& AL & 0.0011 & 0.1048 & -0.55(-0.03\%) & 0.0011 & 0.1174 & 41.69(2.57\%) \\
& VBL & 0.0015 & 0.0995 & -80.49(-4.95\%) & 0.0043 & 0.1026 & 226.11(13.91\%) \\
\midrule
\multirow{3}{*}{HGT} 
& MSE & 0.0009 & 0.0135 & -3.19(-0.2\%) & 0.0009 & 0.0223 & -4.34(-0.27\%)\\
& AL & 0.0009 & 0.0249 & -3.39(-0.21\%) & 0.0011 & 0.0249 & 10.11(0.62\%) \\
& VBL & 0.0009 & 0.0270 & 1.30(0.08\%) & 0.0010 & 0.0249 & -9.81(-0.6\%) \\
\bottomrule
\end{tabular}%
}
\end{table}

\begin{table}[!htbp]
\centering
\scriptsize
\caption{Evaluation on case57 from single- and multi-topology training with selective models and three different loss functions.}
\label{tab:constraint_violations_case57}
\resizebox{\columnwidth}{!}{%
\begin{tabular}{llccc|ccc}
\toprule
\multicolumn{2}{c}{\textbf{Trained on}} & \multicolumn{3}{c}{\textbf{Single (case57)}} & \multicolumn{3}{c}{\textbf{Multi (30+57+118)}} \\
\cmidrule(lr){3-5} \cmidrule(lr){6-8}
\textbf{Arch.} & \textbf{Loss} & \textbf{OPF Sol. Err. } & \textbf{Viol.} & \textbf{Cost Diff.} & \textbf{OPF Sol. Err. } & \textbf{Viol.} & \textbf{Cost Diff.} \\
\midrule
\multirow{3}{*}{GAT} 
& MSE & 0.1263 & 2.620 & -7526.33(-15.47\%) & 0.0143 & 0.5341 & 274.15(0.56\%) \\
& AL & 0.0209 & 0.0718 & -1969.01(-4.05\%) & 0.0123 & 0.0718 & -661.08(-1.36\%) \\
& VBL & 0.2437 & 2.8357 & -11723.59(-24.10\%) & 0.5510 & 4.5343 & 14764.36(30.35\%) \\
\midrule
\multirow{3}{*}{Trans.} 
& MSE & 0.0103 & 0.0565 & -48.16(-0.10\%) & 0.0263 & 0.1926 & 498.63(1.03\%) \\
& AL & 0.0126 & 0.0718 & 592.03(1.22\%) & 0.0262 & 0.0718 & 656.70(1.35\%) \\
& VBL & 0.0252 & 0.0718 & 53.67(0.11\%) & 0.0260 & 0.0718 & 401.57(0.83\%) \\
\midrule
\multirow{3}{*}{HGT} 
& MSE & 0.0244 & 0.0684 & -136.02(-0.28\%) & 0.0252 & 0.0745 & -192.29(-0.39\%) \\
& AL & 0.0094 & 0.018 & -221.39(-0.45\%) & 0.0092 & 0.018 & -117.43(-0.24\%) \\
& VBL & 0.0247 & 0.018 & -150.33(-0.31\%) & 0.0094 & 0.018 & 186.50(0.38\%) \\
\bottomrule
\end{tabular}%
}
\end{table}

\begin{table}[!htbp]
\centering
\scriptsize
\caption{Evaluation on case118 from single- and multi-topology training with selective models and three different loss functions.}
\label{tab:constraint_violations_case118}
\resizebox{\columnwidth}{!}{%
\begin{tabular}{llccc|ccc}
\toprule
\multicolumn{2}{c}{\textbf{Trained on}} & \multicolumn{3}{c}{\textbf{Single (case118)}} & \multicolumn{3}{c}{\textbf{Multi (30+57+118)}} \\
\cmidrule(lr){3-5} \cmidrule(lr){6-8}
\textbf{Arch.} & \textbf{Loss} & \textbf{OPF Sol. Err. } & \textbf{Viol.} & \textbf{Cost Diff.} & \textbf{OPF Sol. Err. } & \textbf{Viol.} & \textbf{Cost Diff.} \\
\midrule
\multirow{3}{*}{GAT} 
& MSE & 0.243 & 22.34 & -40301.82(-26.33\%) & 1.477 & 65.92 & -64212.43(-41.96\%) \\
& AL & 0.1334 & 0.0488 & -16940.96(-11.07\%) & 0.1202 & 0.0531 & -11063.21(-7.23\%) \\
& VBL & 1.699 & 66.49 & -108897.34(-71.16\%) & 1.9389 & 5.5245 & -99349.01(-64.92\%) \\
\midrule
\multirow{3}{*}{Trans.} 
& MSE & 0.0114 & 0.1136 & -224.07(-0.15\%) & 0.1097 & 2.5862 & -16741.81(-10.94\%) \\
& AL & 0.0120 & 0.0487 & 656.37 (0.43\%) & 0.0182 & 0.0487 & -7827.14(-5.11\%) \\
& VBL & 0.0134 & 0.2895 & -4901.14(-3.20\%) & 0.0308 & 0.1241 & -9561.60(-6.25\%) \\
\midrule
\multirow{3}{*}{HGT} 
& MSE & 0.0111 & 0.0687 & -297.97(-0.19\%) & 0.0112 & 0.4872 & 334.86(0.22\%) \\
& AL & 0.0111 & 0.0122 & -297.71(-0.19\%) & 0.0241 & 0.0124 & -7350.06(-4.80\%) \\
& VBL & 0.0111 & 0.0122 & -295.25(-0.19\%) & 0.0115 & 0.1824 & -850.18(-0.56\%) \\
\bottomrule
\end{tabular}%
}
\end{table}

\begin{figure}[!htbp]
    \centering
    \includegraphics[width=0.5\linewidth]{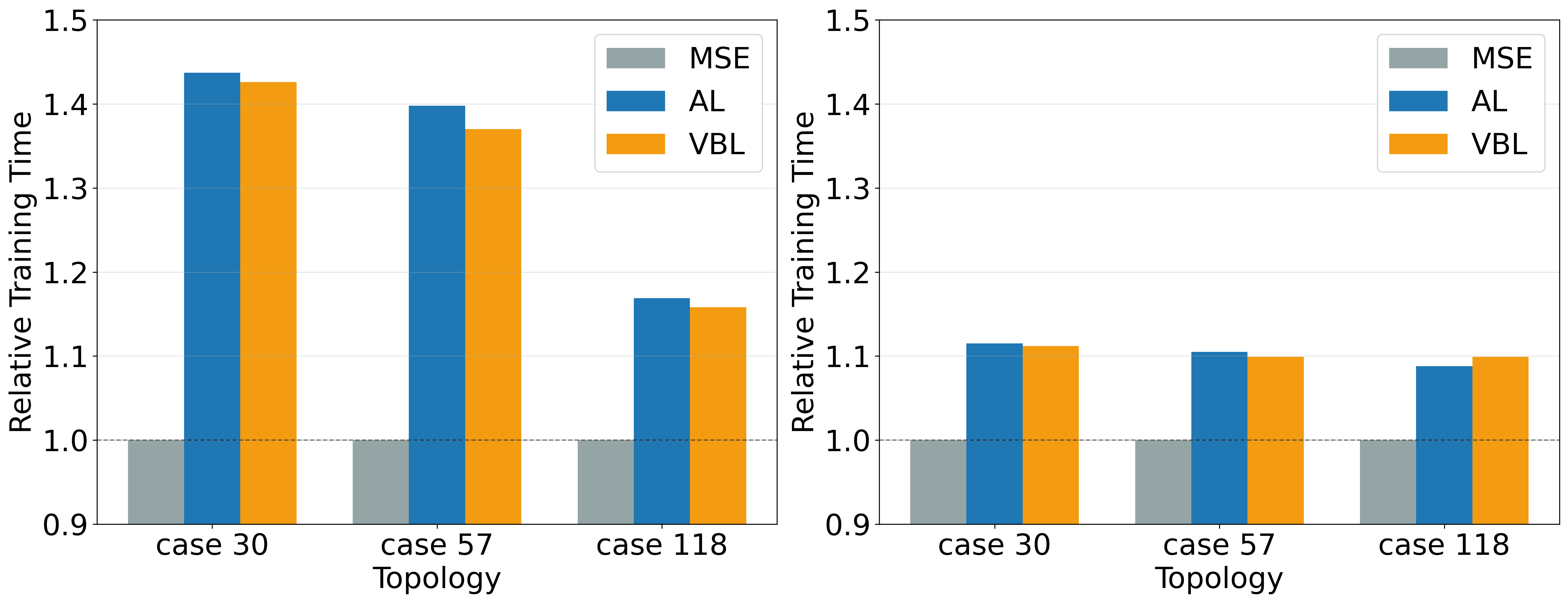}
    \caption{Relative training time for Transformer (left panel) and HGT (right panel) with AL and VBL loss functions normalized to MSE baseline across different topologies.}
    \label{fig:relative_training_time}
\end{figure}

Figure~\ref{fig:relative_training_time} shows the relative training time for Transformer (left panel) and HGT (right panel) with AL and VBL loss functions normalized to MSE baseline across different topologies of increasing size. The Transformer exhibits decreasing relative overhead as topology scales, suggesting that complex loss computations become relatively less expensive compared to the attention mechanism. In contrast, HGT maintains consistently low overhead (approximately 10-11\%) across all topologies, demonstrating that its sparse message-passing architecture makes loss function complexity nearly negligible regardless of system size. 

\subsection{Scaling Behavior Across Network Sizes.}
To assess model scalability, we train Transformer with MSE loss function on topologies of increasing size (case30 through case2000) using a fixed computational budget of 1M training samples per topology. Table~\ref{tab:scaling_cases} reports performance across five network scales spanning two orders of magnitude (30 to 2000 buses). Despite the fixed sample budget, the model achieves consistently low OPF solution error across all scales (0.001-0.018), demonstrating that solution quality scales well with network complexity, though additional samples would likely further improve performances on larger networks.

\begin{table}[!htbp]
\centering
\caption{Performance when scaling across case sizes on 1M samples.}
\label{tab:scaling_cases}
\small
\begin{tabular}{lcc}
\toprule
\textbf{Topology} & \textbf{OPF Sol. Err.} $\downarrow$ & \textbf{Viol.} $\downarrow$ \\
\midrule
case30 & \textbf{0.001146} & \textbf{0.0747} \\
case57 & 0.018954 & 0.1744\\
case118 & 0.013262 & 0.6894 \\
case 500 & 0.006156 & 15.3711\\
case 2000 & 0.007079 & 15.6446\\
\bottomrule
\end{tabular}%
\end{table}

\section{Representational Evidence}
\label{app:repr_evidence}

This appendix provides complementary diagnostic evidence on how different architectures and training objectives shape intermediate representations. We focus on two questions suggested by the main results: (i) whether models without explicit typing (e.g., Transformer) nevertheless learn representations that separate grid component roles, and (ii) whether constraint-aware objectives (AL) change the geometry of learned features relative to pure supervised training (MSE). These analyses are descriptive and are not used for model selection.

\subsection{Methodology: PCA visualizations and linear probing}
\label{app:repr_method}

For a trained model, we extract node-level activations $h_i^{(\ell)}$ at selected backbone layers $\ell$ on held-out samples and perform PCA to visualize dominant directions of variation. Unless otherwise specified, PCA is fit on a random subset of activations and applied to a disjoint subset for visualization. We additionally fit simple linear probes (ridge regression for continuous targets; logistic regression for categorical targets) to quantify linear separability of specific grid characteristics from intermediate activations. Probe performance is reported on held-out samples using standard metrics (e.g., $R^2$ for regression). Importantly, probes are trained \emph{post hoc} on frozen activations and do not affect the underlying model.

\subsection{Implicit separation of node roles}
\label{app:implicit_node_types}

Figure 12 visualizes PCA projections of intermediate activations for samples from case118, colored by node type (bus vs.\ generator vs.\ load vs.\ shunt) for representative architectures: GCN (homogeneous message passing), Transformer (homogeneous attention), and HGT (heterogeneous attention with explicit typing). Two qualitative patterns are consistent with the main-paper findings.

First, HGT yields clear separation between component roles in intermediate layers, as expected given type-aware parametrization. Second, the Transformer exhibits visibly stronger clustering by node type than purely message-passing homogeneous baselines (GCN), suggesting that attention-based aggregation can learn intermediate features that reflect component roles even without explicit type-specific weights. This behavior is consistent with Transformer’s comparatively robust performance under topology shift in the main evaluation: representations that distinguish functional roles can help preserve feasibility-relevant patterns across networks.

\begin{figure}[!htpb]
    \centering
    \includegraphics[width=0.8\linewidth]{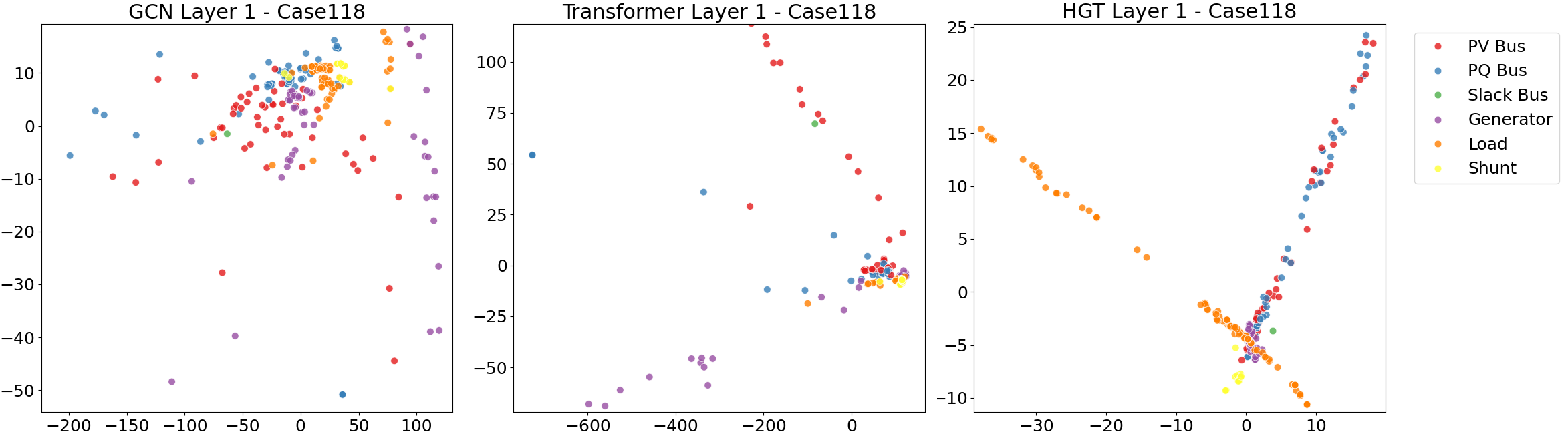}
    \label{fig:implicit_node_types}
    \caption{PCA of layer activations for samples from case118, labeled by node type, for GCN, Transformer, and HGT. The transformer model appears to implicitly encode node type information more similarly to a heterogeneous model irrespective of input types, as seen by the linear separability of node types in the model's intermediate layers.}
\end{figure}

\subsection{Effect of constraint-aware training on representation geometry}
\label{app:al_induced_geometry}

We next compare representations learned under the AL objective versus pure supervised MSE training for the same backbone (HGT). Figure 13 shows PCA projections of activations from the top convolution/attention layer for case30, with points colored by system load. Both objectives produce representations that correlate with load, indicating that the model encodes physically meaningful operating conditions. However, the AL-trained model exhibits a more curved and multi-modal structure in the projected space, while the MSE-trained model appears more linearly organized.

This qualitative difference is consistent with the role of AL: by directly penalizing feasibility residuals, AL can encourage feature transformations that separate regimes where constraints become active (e.g., near operational limits), which may not be well captured by purely accuracy-driven objectives. We emphasize that PCA is a coarse projection; the key takeaway is that constraint-aware training modifies representation geometry in ways aligned with feasibility objectives.

\begin{figure}[!htpb]
\label{fig:al_induced_nonlinearity}
    \centering
    \includegraphics[width=0.45\linewidth]{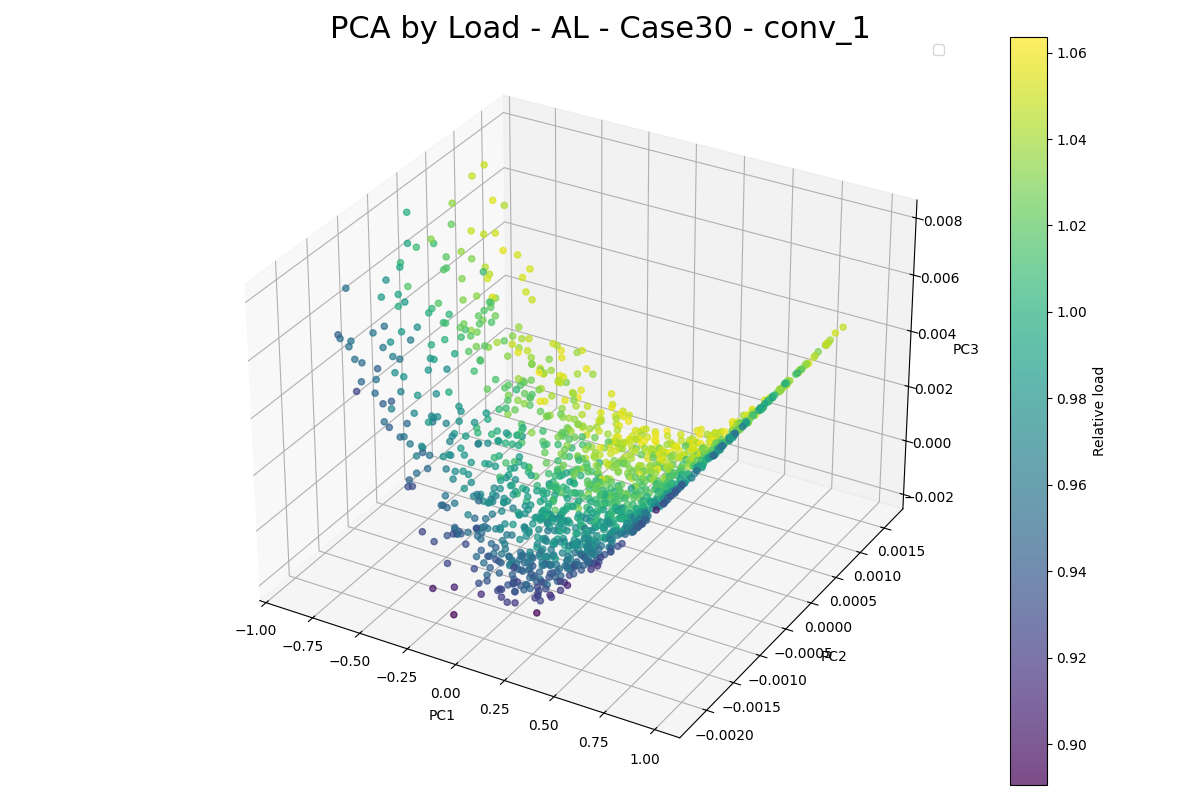}
    \includegraphics[width=0.45\linewidth]{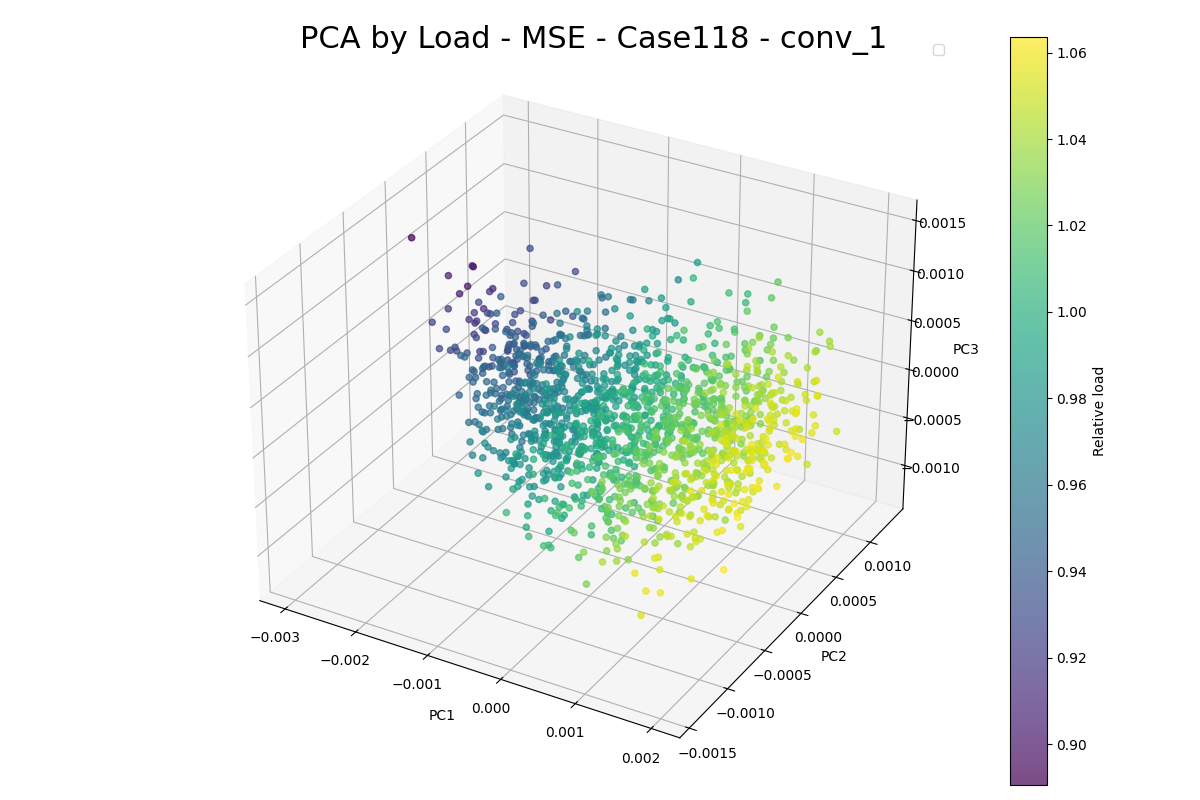}    
    \caption{PCA components of activation for the top layer of convolutions in HGT trained on AL (left) vs MSE (right) losses. We see that both losses lead to the model capturing physical system load, but AL enforces a much more non-linear structure on the model's internal representation.}
\end{figure}

\subsection{Layerwise linear probing: AL increases nonlinearity with depth}
\label{app:al_linear_probe}

\begin{figure}[!htpb]
    \centering
    \includegraphics[width=0.5\linewidth]{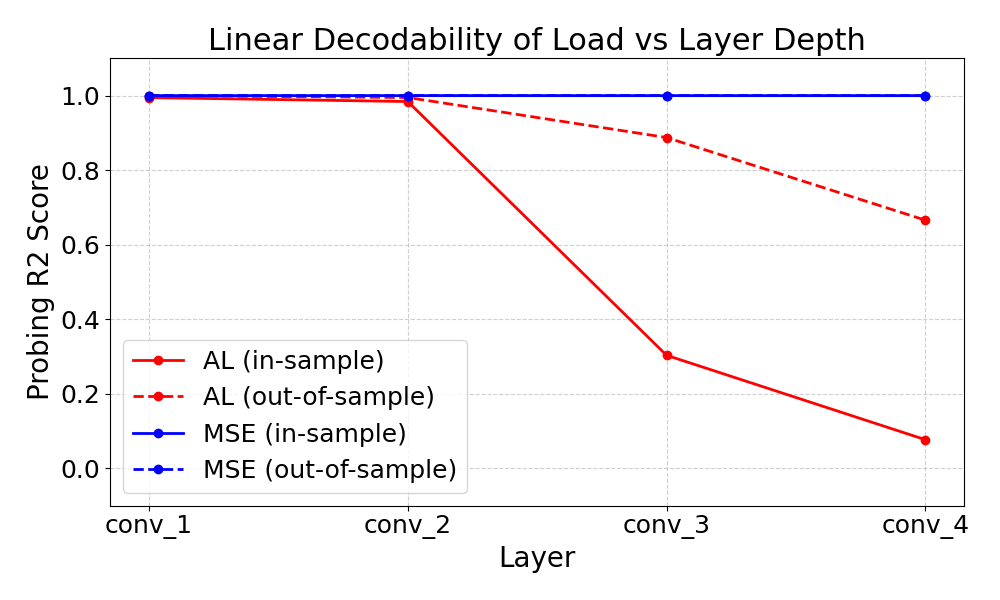}
    \label{fig:al_linear_probing}
    \caption{Linear probing of HGT layers with respect to system load. The choice of AL loss induces additional nonlinearity in the model activations without a decrease in model performance, suggesting that it encourages more abstract feature extraction by the underlying transformer as we progress through network layers.}
\end{figure}

To quantify the preceding observation, we perform layerwise linear probing of intermediate activations with respect to system load. Figure 14 reports probe performance across HGT layers for models trained with AL and MSE. The trends suggest that AL leads to intermediate representations that become less well-explained by a linear map to load as depth increases, while still maintaining overall predictive performance in the main task. A plausible interpretation is that AL encourages progressively more abstract and nonlinear feature extraction, potentially reflecting the need to represent feasibility-relevant regimes beyond a simple linear encoding of operating conditions.

We treat this analysis as supportive evidence rather than a primary claim: linear probe metrics depend on probe capacity and the specific target variable. Nevertheless, the consistent separation between AL and MSE across layers aligns with the empirical improvements in constraint satisfaction reported in the main paper.

\end{document}